\documentclass[sn-mathphys,Numbered]{sn-jnl}

\usepackage{booktabs}\let\cline\cmidrule
\usepackage{graphicx}%
\usepackage{tabularray}
\usepackage{tabularx}
\usepackage{array}
\usepackage{multirow}%
\usepackage{amsmath,amssymb,amsfonts}%
\usepackage{amsthm}%
\usepackage{mathrsfs}%
\usepackage[title]{appendix}%
\usepackage{xcolor}%
\usepackage{textcomp}%
\usepackage{manyfoot}%
\usepackage{booktabs}%
\usepackage{algorithm}%
\usepackage{algorithmicx}%
\usepackage{algpseudocode}%
\usepackage{listings}%
\usepackage{subcaption,siunitx,booktabs}
\usepackage{ragged2e}
\usepackage[normalem]{ulem}
\usepackage[english]{babel}
\usepackage[autostyle, english = american]{csquotes}
\MakeOuterQuote{"}

\usepackage{caption}

\usepackage[export]{adjustbox}

\theoremstyle{thmstyleone}%

\theoremstyle{thmstyletwo}%

\theoremstyle{thmstylethree}%

\begin{document}

\title[]{FAIIR: Building Toward a Conversational AI Agent Assistant for Youth Mental Health Service Provision}

\author*[1,2]{\fnm{Stephen} \sur{Obadinma}}\email{16sco@queensu.ca}

\author[]{\fnm{Alia} \sur{Lachana}}\email{alialachana@gmail.com}
\equalcont{These authors contributed equally to this work.}

\author[2,3]{\fnm{Maia} \sur{Norman}}\email{maia.norman@vectorinstitute.ai}
\equalcont{These authors contributed equally to this work.}

\author[4]{\fnm{Jocelyn} \sur{Rankin}}\email{jocelyn.rankin@gmail.com}
\equalcont{These authors contributed equally to this work.}

\author[2]{\fnm{Joanna} \sur{ Yu}}\email{joanna.yee.yu@gmail.com}
\equalcont{These authors contributed equally to this work.}

\author[1,2]{\fnm{Xiaodan} \sur{Zhu}}\email{xiaodan.zhu@queensu.ca}

\author[4]{\fnm{Darren} \sur{Mastropaolo}}\email{darren.mastropaolo@kidshelpphone.ca}

\author[2]{\fnm{Deval} \sur{Pandya}}\email{deval.pandya@vectorinstitute.ai}

\author[2, 5]{\fnm{Roxana} \sur{Sultan}}\email{roxana.sultan@vectorinstitute.ai}

\author[2,5,6]{\fnm{Elham} \sur{Dolatabadi}}\email{edolatab@yorku.ca}

\affil*[1]{\orgdiv{Electrical and Computer Engineering}, \orgname{Queen’s University}, \orgaddress{\street{99 University Ave}, \city{Kingston},\state{ON}, \country{Canada}}}

\affil[2]{\orgname{Vector Institute}, \orgaddress{\street{W1140-108 College Street, Schwartz Reisman Innovation Campus}, \city{Toronto}, \state{ON}, \country{Canada}}}

\affil[3]{\orgname{University of Waterloo}, \orgaddress{\street{200 University Ave W}, \city{Waterloo},  \state{ON}, \country{Canada}}}

\affil[4]{\orgname{Kids Help Phone}, \orgaddress{\street{439 University Avenue},\city{Toronto},\state{ON},{\country{Canada}}}}

\affil[5]{\orgname{University of Toronto}, \orgaddress{\street{27 King's College Cir}, \city{Toronto},  \state{ON}, \country{Canada}}}

\affil[6]{\orgname{York University}, \orgaddress{\street{4700 Keele St}, \city{Toronto},  \state{ON}, \country{Canada}}}

\abstract{The world’s healthcare systems and mental health agencies face both a growing demand for youth mental health services, alongside a simultaneous challenge of limited resources. Here, we focus on frontline crisis support, where Crisis Responders (CRs) engage in conversations for youth mental health support and assign an issue tag to each conversation. In this study, we develop FAIIR (Frontline Assistant: Issue Identification and Recommendation), an advanced tool leveraging an ensemble of domain-adapted and fine-tuned transformer models trained on a large conversational dataset comprising 780,000 conversations. The primary aim is to reduce the cognitive burden on CRs, enhance the accuracy of issue identification, and streamline post-conversation administrative tasks. We evaluate FAIIR on both retrospective and prospective conversations, emphasizing human-in-the-loop design with active CR engagement for model refinement, consensus-building, and overall assessment. Our results indicate that FAIIR achieves an average AUC ROC of 94\%, a sample average F1-score of 64\%, and a sample average recall score of 81\% on the retrospective test set. We also demonstrate the robustness and generalizability of the FAIIR tool during the silent testing phase, with less than a 2\% drop in all performance metrics. Notably, CRs’ responses exhibited an overall agreement of 90.9\% with FAIIR's predictions. Furthermore, expert agreement with FAIIR surpassed their agreement with the original labels. To conclude, our findings indicate that assisting with the identification of issues of relevance helps reduce the burden on CRs, ensuring that appropriate resources can be provided and that active rescues and mandatory reporting can take place in critical situations requiring immediate de-escalation.}


\maketitle

\section{Introduction}\label{sec1}
Globally, one in seven young individuals aged 10 to 19 years old experience a mental health condition, making a significant contribution to the global burden of disability and disease \citep{world_2021}. Suicide ranks as the fourth leading cause of death among 15 to 29 year olds, and in Canada, one in five individuals will experience a mental illness by age 25 \citep{world_2021,mhccStats}. Despite 70\% of mental illness starting during childhood or adolescence, only a fraction of young individuals are able to access appropriate care, leading to increasing rates of youth hospitalizations for mental health disorders \citep{Wiens2020AGN,mhccStats, cihi}.

Crisis support conversations are a critical component of mental health services, providing immediate, accessible, and empathetic support to youth in distress \citep{gould2007evaluation,hoffberg2020effectiveness}. This approach offers significant benefits, including early intervention, reduced healthcare burdens, and the potential to de-escalate life-threatening situations. However, several challenges persist, including the high demand for these services which is particularly evident in the volume of text conversations received by Kids Help Phone (KHP), a Canadian not-for-profit e-mental health organization. Since the launch of its text service in 2018, KHP has facilitated over 1 million Short Message Service (SMS) interactions, with a significant 51\% increase observed during the COVID-19 pandemic in 2020. This heightened demand underscores the necessity of expanding the team of Crisis Responders (CRs), composed of both professionals and trained volunteers. The complexity of the issues discussed by youth during these conversations and the cognitive burden on CRs managing emotionally stressed individuals in potentially life-critical situations \citep{dinakar2015mixed} add further challenges to crisis support services. Processing nuances in youth natural language from diverse populations and intersectionalities is not straightforward. Moreover, CRs must complete post-conversation surveys to identify key issues such as suicide and abuse, further adding to their workload and time commitment. 

In this study, we focus on building an efficient and scalable support mechanism leveraging Natural Language Processing (NLP) approaches to assist CRs. We utilize one of the largest crisis support conversational datasets, comprising 780,000 interactions by KHP. Our aim is to reduce CR's cognitive burden, improve the accuracy of issue identification, and streamline post-conversation administrative tasks. Dramatic recent improvements in the performance and availability of state-of-the-art large language models (LLMs) \citep{devlin2018bert,Beltagy2020LongformerTL,Touvron2023Llama2O} have accelerated development and application of NLP tools to augment health services, alongside human experts \citep{info:doi/10.2196/15708, dolatabadi2023using,raza2023discovering,dolatabadi2023natural}. In the mental health context, NLP tools have been developed to identify signs of depression \citep{healthcare10020291,hu2021bluememo,yates2017depression} and suicidal intentions \citep{sinha2019suicidal} within social media posts. In addition, they can identify themes related to suicide \citep{downs2017detection} and mental health status \citep{tran2017predicting} from clinical notes. These tools are also utilized to aid in triage and reduce wait times on message-based suicide support platforms \citep{SwaminathanDetectionMentalHealth}. These real-world applications highlight NLP's potential in augmenting mental health support across both clinical and non-clinical communication channels.

To this end, we developed a human-centric classification and information retrieval model called "Frontline Assistant: Issue Identification and Recommendation" (FAIIR). The FAIIR tool inherits intent identification outlined in the Neural Agent Assistant framework, designed to improve AI-enabled conversational tasks \citep{obadinma-etal-2022-bringing}, as well as state-of-the-art transformer-based models built for long conversations \citep{Li2022ClinicalLongformerAC}. The FAIIR development process first included optimizing and fine-tuning a suite of transformer models that classify conversations into a predefined set of 19 clinically-orientated care paradigms, referred to as issue tags, such as suicidality and abuse (refer to tags defined in Appendix \ref{secA00}). Following this, the process encompassed ensemble techniques and human-in-the-loop validation with active engagement of CRs. CRs' involvement was crucial for validation, consensus-building, refining performance, and ensuring the tool's utility. To uphold responsible AI development standards, we conducted thorough evaluations for bias and fairness. Finally, we presented the outcomes of our initial silent trial, detailed our implementation plans, and explored additional potential use cases for the FAIIR tool.

\section{Results}
\subsection{FAIIR Model Development and Issue Tag Prediction in Crisis Conversations}\label{sec2}

\begin{figure*}[!htbp]
\centering
\includegraphics[width=0.99\textwidth, center]{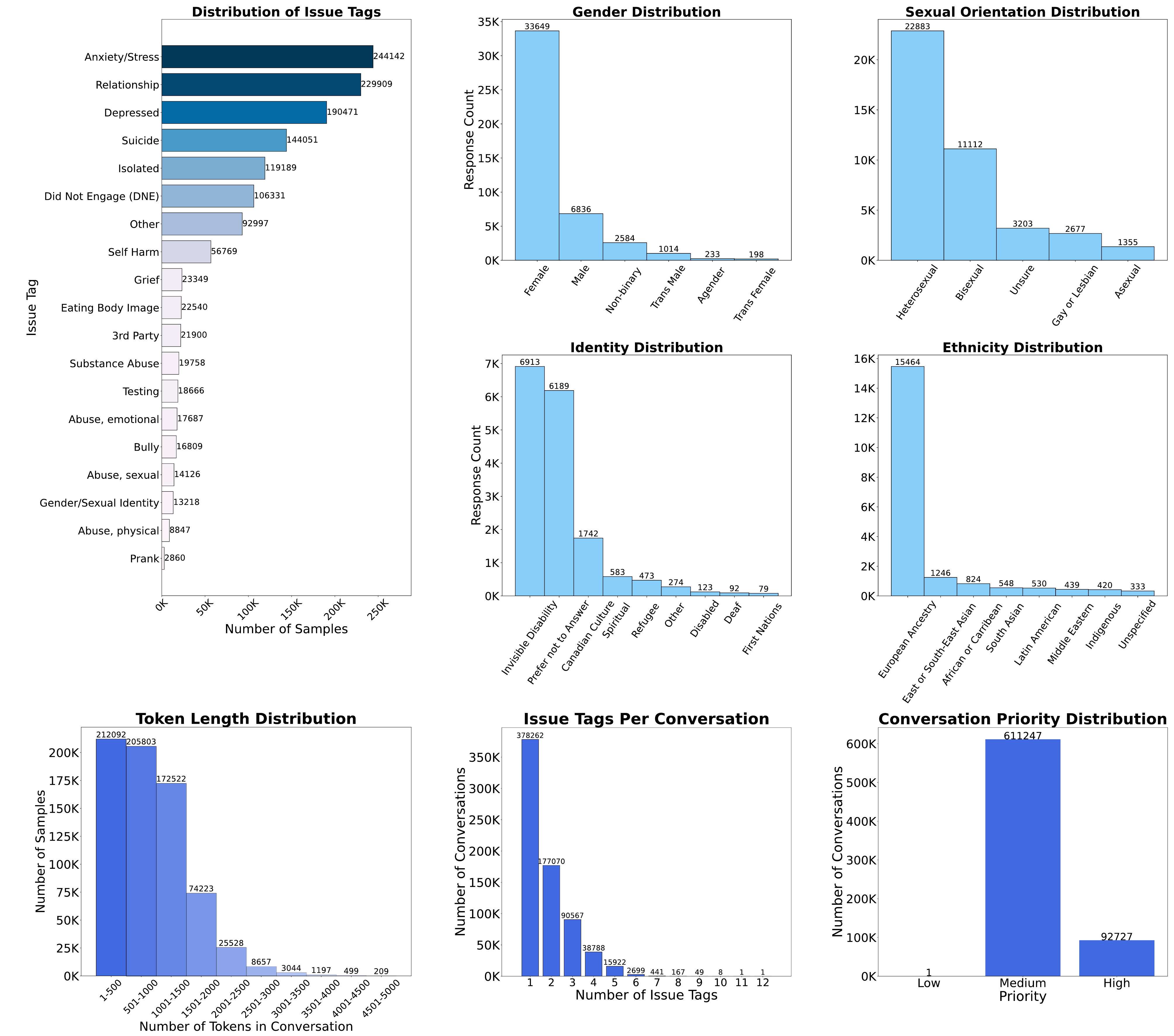}
\caption{ \textit{Dataset Statistics}: \textbf{(Top-Left)} 703,975 youth conversations with frontline crisis responders are classified into 19 pre-defined issue tags. Multiple tags may be assigned per conversation, as relevant. \textbf{(Top-Middle to Middle-Right)} After each interaction, service users are invited to complete a demographic survey, gauging the conversation's helpfulness and the individual's demographics (age range, ethnicity, and identification with specific identity groups). The distribution of each demographic category of the aggregated surveys is presented. \textbf{(Bottom-Left)} Distribution of conversation lengths (\# tokens). \textbf{(Bottom-Middle)} Distribution of the number of issue tags assigned per conversation across the dataset. \textbf{(Bottom-Right)} Distribution of priority labels assigned to conversations. } 
\label{fig:figure1} 
\end{figure*}

\noindent  \textbf{Crisis Support Conversational Dataset}\\
To develop FAIIR, we leveraged 703,975 anonymized text-based crisis support conversations exchanged between service users and CRs from January 2018 to February 2023. Each conversation consists of multiple message exchanges (multi-turn dialogues), making it essential for the model to process long-form text effectively. Identifiable information was scrubbed to ensure privacy compliance. Conversations vary significantly in length in terms of tokens. Tokens in this context refer to discrete units of text that typically represent words, sub-words, characters, or punctuation obtained after breaking a sequence of text down so that it can be processed by a model. The average and median number of tokens per conversation were 913 and 850 respectively (Figure \ref{fig:figure1}) with the majority of conversations (53\%) between 500 and 1,500 tokens, and only a small number extending above 3,000 (0.7\%). Given this distribution, we set a 2,000-token maximum input length, covering 94.4\% of all conversations (Figure \ref{fig:figure1}). An optional demographic survey was completed by 17\% of service users (59,603 survey respondents of a total of n=340,512 overall service users), with responses originating mostly from individuals identifying as female (75\%), heterosexual (55\%), and of European ancestry (77.5\%) as shown in Figure \ref{fig:figure1}. It is essential to note that only a small number of overall service users completed the survey, and thus the results do not fully represent the distribution or demographic features of service users overall.\\\\

\noindent  \textbf{Issue Tag Prediction}\\
FAIIR is a multi-label classification model that classifies conversations into 19 predefined issue categories, with it being possible for multiple issue categories applying for a given conversation. However, the distribution of issue tags in the data is highly imbalanced (Figure \ref{fig:figure1}), with the most frequent tag, \textit{Anxiety/Stress}, appearing in over 244,000 conversations, and the least frequent, \textit{Prank}, appearing in only 2,800 conversations. Additionally, issue tagging varies across conversations: 53.73\% of conversations are assigned a single issue tag, while 46\% contain between 2 and 9 issue tags, reflecting the complexity of the challenges faced by service users.
Regarding risk levels, the majority of conversations were classified as high-risk (13\%) or medium-risk (87\%), based on the methodology detailed in Appendix \ref{secA3}.

  \vspace{6mm}

\begin{table*}[!htbp]
\caption{The fine-grained performance of the FAIIR tool, an ensemble of three Longformer models, is reported for two datasets: a retrospective test set (n=140,795) and a prospective silent testing set (n=84,832). Each model was pre-trained and fine-tuned to identify multiple issue tags within conversations using the full training and validation set of 563,180 conversations. Results are provided as averages across all samples for classification thresholds of 0.5, 0.25, and an updated threshold (used only for the prospective silent testing dataset).}

\label{table:mainresultslongformer}
\centering
\resizebox{\textwidth}{!}{
\begin{tabular}{llccccccccccccccccccc}
\toprule
\multicolumn{2}{l}{}& {\rotatebox{90}{3rd Party}} & {\rotatebox{90}{Abuse, Emotional}} & {\rotatebox{90}{Abuse, Physical}} & {\rotatebox{90}{Abuse, Sexual}} & {\rotatebox{90}{Anxiety/Stress}} & {\rotatebox{90}{Bully}} & {\rotatebox{90}{Depressed}} & {\rotatebox{90}{DNE}}  & {\rotatebox{90}{Eating Body Image}} & {\rotatebox{90}{Gender/Sexual Identity}} & {\rotatebox{90}{Grief}} & {\rotatebox{90}{Isolated}} & \rotatebox{90}{Other} & \rotatebox{90}{Prank} & \rotatebox{90}{Relationship} & \rotatebox{90}{Self Harm} & \rotatebox{90}{Substance Abuse} & \rotatebox{90}{Suicide} & \rotatebox{90}{Testing} \\ \midrule
\multicolumn{21}{c}{\textbf{Test set}}\\ \midrule
& AUC ROC & 0.99 & 0.96 & 0.98 & 0.99 & 0.87 & 0.96 & 0.96 & 0.85 & 0.98 & 0.99 & 0.95 & 0.87 & 0.74 & 0.96 & 0.9 & 0.97 & 0.97 & 0.95 & 0.97 \\  \midrule
\multirow{3}{*}{Threshold - 0.25} & Precision & 0.64 & 0.35 & 0.33 & 0.47 & 0.61 & 0.3 & 0.66 & 0.51 & 0.51 & 0.52 & 0.34 & 0.47 & 0.28 & 0.34 & 0.63 & 0.55 & 0.32 & 0.63 & 0.41 \\
 & Recall & 0.95 & 0.7 & 0.79 & 0.91 & 0.81 & 0.79 & 0.86 & 0.81 & 0.88 & 0.92 & 0.83 & 0.7 & 0.45 & 0.66 & 0.85 & 0.92 & 0.86 & 0.87 & 0.73 \\
 & F1 & 0.76 & 0.46 & 0.47 & 0.62 & 0.69 & 0.43 & 0.75 & 0.62 & 0.64 & 0.67 & 0.48 & 0.56 & 0.35 & 0.45 & 0.73 & 0.69 & 0.46 & 0.73 & 0.53 \\ \midrule
\multirow{3}{*}{Threshold - 0.5} & Precision & 0.69 & 0.53 & 0.51 & 0.57 & 0.79 & 0.46 & 0.79 & 0.67 & 0.64 & 0.6 & 0.45 & 0.66 & 0.4 & 0.47 & 0.77 & 0.64 & 0.42 & 0.75 & 0.61 \\
 & Recall & 0.92 & 0.5 & 0.63 & 0.82 & 0.56 & 0.66 & 0.75 & 0.51 & 0.79 & 0.87 & 0.75 & 0.4 & 0.23 & 0.55 & 0.66 & 0.84 & 0.75 & 0.73 & 0.59 \\
 & F1 & 0.79 & 0.51 & 0.56 & 0.67 & 0.66 & 0.54 & 0.77 & 0.58 & 0.71 & 0.71 & 0.56 & 0.5 & 0.29 & 0.51 & 0.71 & 0.73 & 0.54 & 0.74 & 0.6  \\ \midrule
\multicolumn{21}{c}{\textbf{Silent Testing}} \\ \midrule
 & AUC ROC & 0.99 & 0.96 & 0.99 & 0.99 & 0.86 & 0.97 & 0.95 & 0.85 & 0.98 & 0.98 & 0.95 & 0.87 & 0.73 & 0.97 & 0.9 & 0.97 & 0.97 & 0.94 & 0.95 \\  \midrule
\multirow{3}{*}{Threshold - 0.25} & Precision & 0.58 & 0.45 & 0.48 & 0.52 & 0.76 & 0.43 & 0.8 & 0.61 & 0.58 & 0.48 & 0.44 & 0.62 & 0.43 & 0.37 & 0.75 & 0.63 & 0.44 & 0.72 & 0.48 \\
 & Recall & 0.9 & 0.48 & 0.7 & 0.82 & 0.53 & 0.71 & 0.77 & 0.47 & 0.8 & 0.84 & 0.75 & 0.38 & 0.21 & 0.46 & 0.67 & 0.85 & 0.79 & 0.74 & 0.48 \\
 & F1 & 0.7 & 0.46 & 0.57 & 0.63 & 0.62 & 0.54 & 0.79 & 0.53 & 0.67 & 0.61 & 0.56 & 0.47 & 0.28 & 0.41 & 0.71 & 0.72 & 0.57 & 0.73 & 0.48 \\  \midrule
\multirow{3}{*}{Threshold - 0.5} & Precision & 0.52 & 0.29 & 0.3 & 0.41 & 0.58 & 0.29 & 0.69 & 0.43 & 0.44 & 0.42 & 0.33 & 0.43 & 0.33 & 0.26 & 0.63 & 0.55 & 0.34 & 0.59 & 0.27 \\
 & Recall & 0.94 & 0.69 & 0.84 & 0.91 & 0.79 & 0.86 & 0.86 & 0.78 & 0.88 & 0.89 & 0.82 & 0.67 & 0.43 & 0.57 & 0.86 & 0.92 & 0.87 & 0.87 & 0.62 \\
 & F1 & 0.67 & 0.41 & 0.44 & 0.57 & 0.67 & 0.43 & 0.77 & 0.56 & 0.58 & 0.57 & 0.47 & 0.52 & 0.37 & 0.36 & 0.72 & 0.68 & 0.49 & 0.7 & 0.38 \\  \midrule
\multirow{3}{*}{Threshold - updated}& Precision & 0.51 & 0.26 & 0.26 & 0.39 & 0.7 & 0.25 & 0.67 & 0.53 & 0.4 & 0.4 & 0.3 & 0.48 & 0.3 & 0.25 & 0.71 & 0.52 & 0.31 & 0.62 & 0.24 \\
 & Recall & 0.94 & 0.74 & 0.86 & 0.92 & 0.63 & 0.88 & 0.87 & 0.6 & 0.89 & 0.89 & 0.84 & 0.6 & 0.49 & 0.61 & 0.75 & 0.93 & 0.89 & 0.85 & 0.66 \\
 & F1 & 0.66 & 0.38 & 0.4 & 0.55 & 0.66 & 0.39 & 0.76 & 0.56 & 0.56 & 0.55 & 0.44 & 0.53 & 0.37 & 0.35 & 0.73 & 0.67 & 0.46 & 0.71 & 0.35\\

\bottomrule
\end{tabular}}
\end{table*}

\noindent  \textbf{Ensemble of Longformer Models Excelled in FAIIR Issue Tag Predictions}\\
We first compared four transformer-based LLMs, each fine-tuned on a dataset containing 50,000 randomly sampled conversations (see Appendix \ref{secA1}). Transformer-based LLMs \citep{NIPS2017_3f5ee243} are advanced deep learning models designed to analyze large volumes of text, such as patient-provider conversations, and identify patterns or extract insights. Among the models tested, Longformer demonstrated the best performance, excelling accuracy and ability to effectively handle long conversations without losing contextual meaning. 

Given Longformer's superior performance, the backend of the FAIIR tool was built using an ensemble of three Longformer \citep{Beltagy2020LongformerTL} models which involves combining the predictions of separate models with slightly varied initialization and fine-tuning processes, aiming to enhance overall accuracy and reliability. These models were adapted to the mental health support domain through the technique of masked language modelling, a training process which aids in model understanding of specific language, terminology, and the context of words by leveraging the entire dataset to get the model to learn to predict masked tokens. 

After domain adaptation, the models were further fine-tuned for the multi-label classification task. This task involved training the models to identify multiple relevant issues at hand (see more details about predefined issue tags in Appendix \ref{secA00}). The fine-tuning was conducted using a larger dataset of 563,180 conversations - 422,385 used for training, and 140,795 for validation. The models were then evaluated retrospectively on an independent test set of 140,795 conversations to confirm their performance in a real-world context. 

\begin{figure*}[!htb]
\centering
\includegraphics[width=.49\textwidth]{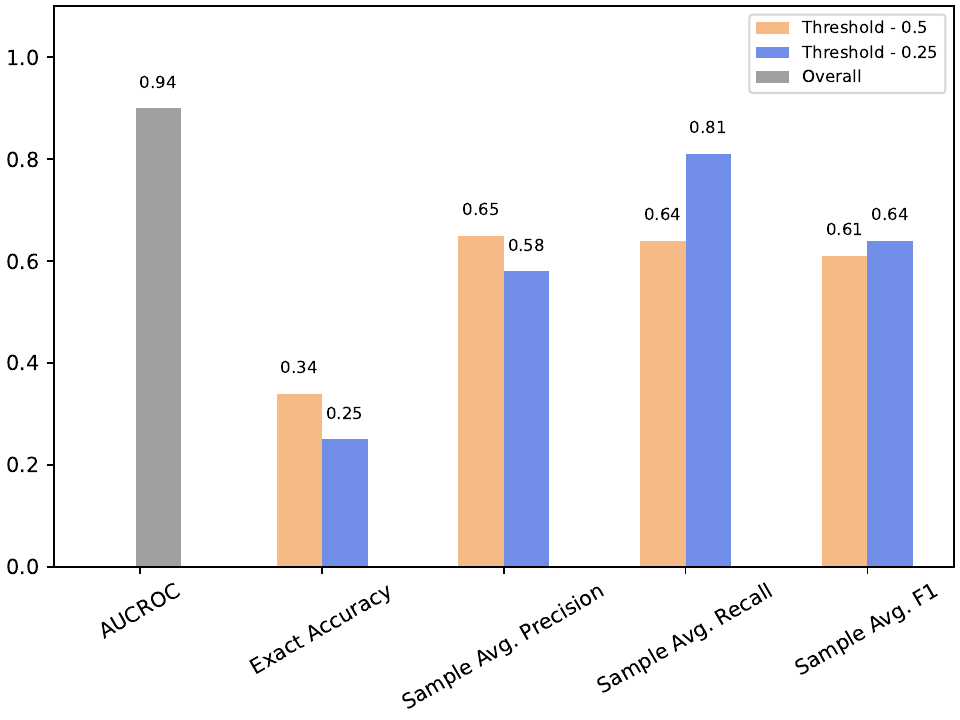}\hspace{0.2em}%
\includegraphics[width=.49\textwidth]{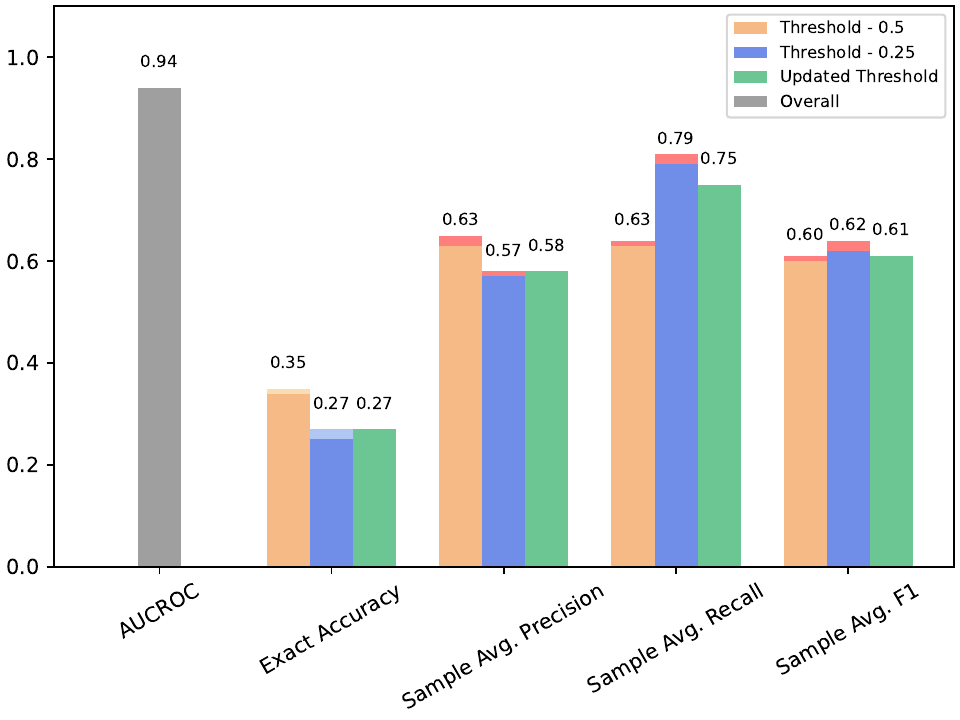}\hspace{0.2em}%
\caption{\textbf{(Left)} Averaged performance of the FAIIR tool in predicting all 19 issue tags is shown for the retrospective test set (n=140,795). \textbf{(Right)} Averaged performance of the FAIIR tool across all issue tags in the silent testing prospective test sets (n=84,932), evaluated using three classification thresholds. For silent testing, results for the silent testing overlay retrospective results, with decreases in performance highlighted in red and gains shown in a lighter shade. The AUC ROC bar represents the average ability of the tool to distinguish between issue tags across all categories. The tool's best overall performance is an F1-score of 0.64 on the retrospective test set and 0.62 on the silent testing prospective test set.}\label{table:LongformerOverallResults}
\end{figure*}

The performance of the ensemble model on the test set, broken down by individual issue tags, is detailed in Table \ref{table:mainresultslongformer}, while the average performance across all issue tags is shown in Figure \ref{table:LongformerOverallResults} (Left). The majority of the Area Under the Receiver Operating Characteristic Curve (AUC ROC) scores exceeded 0.9, indicating strong overall model performance, with the lowest score being 0.74. 

Due to the imbalanced nature of the data, whereby certain issue tags are under- or over-represented, precision (proportion of correct positive predictions) is relatively low ($<$ 0.65). However, recall (proportion of true positives identified correctly) remains high ($<$ 0.9), alongside overall accuracy (0.94). This means that the model is highly effective at identifying relevant tags, but may occasionally also identify irrelevant issues in conversation, especially for less common topics. 

Classification thresholds define the level of confidence a model must have before assigning an issue tag. The model generates 19 confidence scores—one for each issue label—based on the probability that a label applies to a given conversation. A threshold is then used to set the minimum confidence score required to definitively assign a label. When selecting the optimal threshold for our context, a threshold of 0.25 was found to strike the best balance between precision and recall, thereby allowing the model to capture critical issues, even if irrelevant tags are sometimes included. At this threshold, the sample average F1-score (a combined measure of precision and recall) was 0.64, with a sample average recall of 0.81 and a sample average precision of 0.58. The trade-off between recall and precision is acceptable within this context, as it prioritizes capturing critical issues , even if some irrelevant issue tags are also included.

The model performed exceptionally well in identifying high-priority issues such as \textit{Suicide} (F1-score = 0.73), \textit{Depression} (F1-score = 0.75), \textit{Relationship Problems} (F1-score = 0.73), \textit{Self Harm} (F1-score = 0.69), \textit{Anxiety/Stress} (F1-score = 0.69), and \textit{Third-Party} conversations (F1-score = 0.76). However, model performance was relatively poorer for rarer issues, including \textit{Other} (F1-score = 0.35), \textit{Abuse, Emotional} (F1-score = .46), \textit{Abuse, Physical} (F1-score = 0.47), \textit{Isolated} (F1-score = 0.56), \textit{Prank} (F1-score = 0.45), and \textit{Testing} (F1-score = 0.53). These results suggest that while the model is effective at identifying common or critical issues, additional refinement may be required to improve performance across less frequent categories.

\noindent\begin{minipage}{\linewidth}
\centering
\captionof{table}{The performance of FAIIR (with threshold 0.25) within subgroups of service users across four distinct demographic categories. Demographic information was provided by 17\% of service users. }\label{Table:DemographicPerformance}
\resizebox{\linewidth}{!}{%
\begin{tabular}{lccccc} 
\hline
\hline
 & \textbf{Precision} & \textbf{Recall} & \textbf{F1-score} & \textbf{Accuracy} & \textbf{$\#$ Conversations} \\ 
\cline{2-6}
\textbf{Gender} &  &  &  &  &  \\
~ ~Male & 0.56 & 0.86 & 0.64 & 0.90 & 1236 \\
~ ~Female & 0.57 & 0.86 & 0.65 & 0.90 & 5955 \\
~ ~Trans Male & 0.56 & 0.85 & 0.64 & 0.90 & 173 \\
~ ~Trans Female~ & 0.49 & 0.86 & 0.59 & 0.89 & 32 \\
~ ~Non-binary~ & 0.56 & 0.85 & 0.63 & 0.90 & 501 \\
~ ~Agender~ & 0.59 & 0.80 & 0.65 & 0.91 & 35 \\
\hline
\\
\multicolumn{2}{l}{\textbf{Orientation~}} &  &  &  &  \\
~ ~Heterosexual & 0.57 & 0.87 & 0.65 & 0.90 & 4042 \\
~ ~Gay or Lesbian & 0.58 & 0.85 & 0.66 & 0.91 & 516 \\
~ ~Bisexual & 0.56 & 0.85 & 0.64 & 0.90 & 1969 \\
~ ~Asexual & 0.59 & 0.86 & 0.66 & 0.91 & 247 \\
~ ~Unsure & 0.56 & 0.86 & 0.64 & 0.90 & 586 \\
\hline
\\
\textbf{Identity} &  &  &  &  &  \\
~ ~Canadian Culture & 0.61 & 0.84 & 0.67 & 0.91 & 184 \\
~ ~Disabled & 0.55 & 0.91 & 0.63 & 0.90 & 19 \\
~ ~Refugee~ & 0.55 & 0.82 & 0.63 & 0.90 & 96 \\
~ ~Spiritual~ & 0.61 & 0.89 & 0.68 & 0.90 & 76 \\
~ ~Deaf~ & 0.57 & 0.89 & 0.66 & 0.91 & 17 \\
~ ~First Nations~ & 0.62 & 0.73 & 0.66 & 0.92 & 10 \\
~ ~Invisible Disability & 0.56 & 0.85 & 0.64 & 0.90 & 1095 \\
~ ~Other~ & 0.60 & 0.83 & 0.65 & 0.91 & 36 \\
~ ~Prefer not to Answer & 0.56 & 0.85 & 0.64 & 0.90 & 1048 \\

\hline
\\
\textbf{Ethnicity~} &  &  &  &  &  \\
~ ~European Ancestry & 0.57 & 0.85 & 0.65 & 0.90 & 2376 \\
~ ~African or Caribbean~ & 0.51 & 0.85 & 0.61 & 0.90 & 134 \\
~ ~Indigenous~ & 0.56 & 0.88 & 0.65 & 0.89 & 74 \\
~ ~East or South-East Asian~ & 0.55 & 0.86 & 0.63 & 0.90 & 190 \\
~ ~Middle Eastern~ & 0.53 & 0.87 & 0.62 & 0.89 & 73 \\
~ ~Latin American~ & 0.55 & 0.88 & 0.64 & 0.90 & 86 \\
~ ~South Asian~ & 0.56 & 0.85 & 0.64 & 0.90 & 87 \\
~ ~Unspecified & 0.61 & 0.90 & 0.69 & 0.90 & 46 \\
\hline
\hline
\end{tabular}
}
\end{minipage} \vspace{6mm}

\vspace{1mm}

\noindent  \textbf{FAIIR Tool Predictions Observed to be Unbiased Across Demographic Subgroups}\\
 Table \ref{Table:DemographicPerformance} demonstrates the performance (precision, recall, F1-score, and accuracy) of the FAIIR tool in predicting issue tags within 27 distinct subgroups across four demographic categories, representing 17\% of overall service users (See distribution in Figure \ref{fig:figure1}). The standard deviation of F1-scores within each demographic category is less than $0.025$ (for Gender: $\pm0.023$, Orientation: $\pm0.010$, Identity: $\pm0.018$, and Ethnicity: $\pm0.024$), indicating a consistent model performance with a narrow gap across varying demographic subgroups. The results of a one-sample t-test (p-value $<$ 0.001) show that there is no significant difference between the F1-scores of individual demographic subgroups and the overall performance of the model. This indicates that the model performs consistently across different demographic groups, and any differences in scores are unlikely due to sampling bias.

\subsection{Expert Assessment and Evaluation of FAIIR Predictions}
In total, we solicited 240 annotations from 12 experts, with each conversation undergoing independent assessment by three distinct assessors for open review and another three for blind review. The two types of review differed in the presentation of the FAIIR predictions: in the open review (Figure ~\ref{fig:SurveyScreenshot} in Appendix ~\ref{secA4}), assessors were provided with FAIIR’s predicted issue tags as a reference, whereas in the blind review, assessors identified issue tags independently without any exposure to FAIIR’s predictions.\\

\begin{figure*}[!htb]
\centering
\includegraphics[width=.99\textwidth]{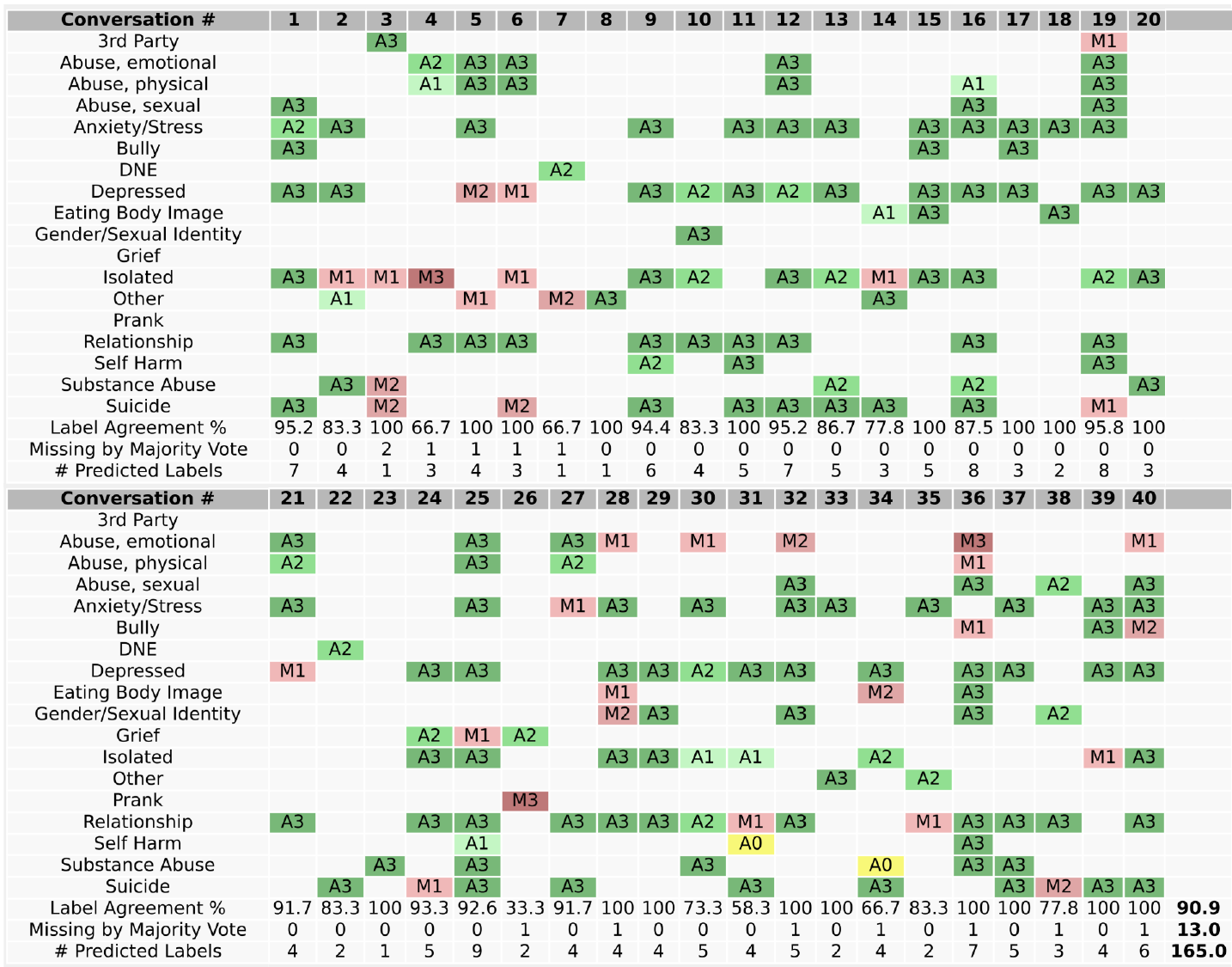}\hspace{0.2em}%
\caption{Experts' blind review results presented in a matrix format, whereby each row represents an issue tag and each column a conversation. Three reviewers assess each conversation, providing feedback on the issue tags predicted by the FAIIR tool: indicating their agreement or disagreement, and identifying missing tags, where applicable. Cells shaded in green indicate agreement between reviewer and model, while cells shaded in red represent missing tags. The letter `A' in the cell followed by a number indicates the total number of reviewers (of three total) in agreement with model predictions. The letter `M' in the cell followed by a number indicates the total number of reviewers who believe this issue tag was missed by the FAIIR tool.}  \label{tablesurveyA}
\end{figure*}

\noindent \textbf{FAIIR's Predictions Validated by Human Experts}\\
For results of the blind assessment demonstrated in Figure \ref{tablesurveyA}, on average across 40 conversations, 90.9\% agreement (lowest: 33\% and highest 100\%) was achieved overall between CRs and the FAIIR tool, where FAIIR was able to predict 165 issue tags and by majority agreement it only missed 13 tags. Among all issue tags, agreement was reached in the majority of instances of \textit{Anxiety/Stress}, \textit{Bully}, \textit{Relationship}, \textit{3rd Party}, \textit{Suicide}, and \textit{Abuse, Emotional}; with more frequent discordance for the issue tags of \textit{Grief}, \textit{Self Harm}, \textit{Abuse, Physical}; \textit{Other}, and\textit{ Eating Body Image}. \\

\begin{figure}[!htb]
\centering
\includegraphics[width=1\textwidth]{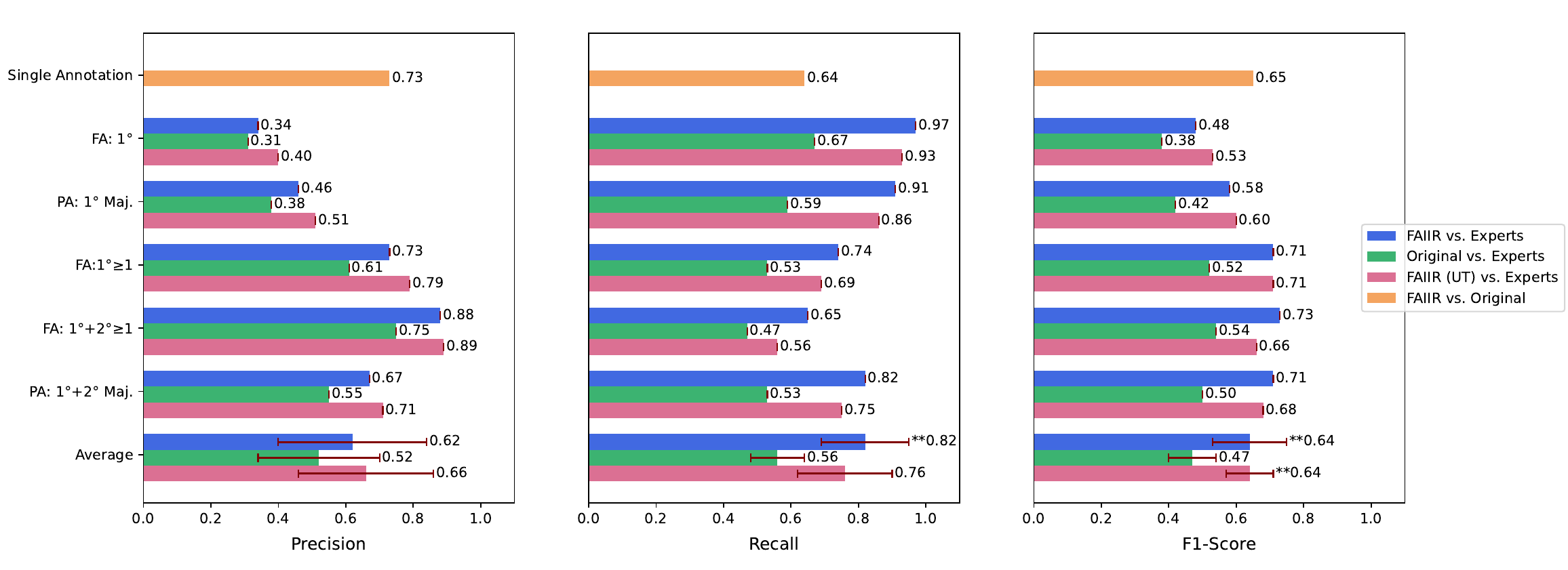}\hspace{0.2em}%
\caption{Comparison of consensus among expert responses, FAIIR tool predictions, and original annotations from open review. Precision, recall, and F1-score measures were averaged across all issue tags and conversations. "FA: 1\(^\circ\)" denotes full agreement on primary issue tags, "PA: 1\(^\circ\) Maj." denotes partial agreement on primary issue tags via majority vote, "PA: 1\(^\circ\)$+$2\(^\circ\) Maj." denotes partial agreement on primary and secondary issue tags via majority vote; "FA: 1 \(^\circ\) \(\geq\) 1" denotes full agreement on primary issue tags via at least one vote; and "FA: 1\(^\circ\)$+$2\(^\circ\) \(\geq\) 1" denotes full agreement on primary and secondary issue tags via at least one vote. "Average" denotes the average performance across all five consensus criteria. One sample t-test was conducted to assess the statistical significance between average and FAIIR tool vs. original annotations (identified by **). The consensus among expert responses and FAIIR predictions after updating the threshold in accordance with expert assessment can be seen in the "FAIIR (UT) vs. Experts" bars.}\label{table:survey_b_consensus}
\end{figure}

\noindent \textbf{FAIIR Understands the Conversational Context}\\
Figure \ref{table:survey_b_consensus} presents a comparison of consensus among the experts' blind responses, FAIIR tool prediction(s), and original issue tag(s)\footnote{Refers to the issue tag(s) recorded in the training dataset, assigned following the original conversation.}. For all consensus comparison settings (Figure \ref{table:survey_b_consensus}), the level of agreement between the FAIIR tool and expert responses (average precision 0.62$\pm$0.22, average recall 0.82$\pm$0.13, and average F1-score 0.64$\pm$0.11; respectively) are higher than original annotations and expert responses (average precision 0.52$\pm$0.18, average recall 0.56$\pm$0.08, and average F1-score 0.47$\pm$0.07; respectively). Thus, when using expert responses as a reference for comparison, FAIIR predictions align more closely with the experts' annotations than the original issue tags. The results of the unpaired t-test on the averaged measures reveal that the concordance between FAIIR and expert responses is significantly higher (p-value $<$ 0.001) compared to the concordance between the original annotations and expert responses. Figure \ref{table:survey_b_consensus} also demonstrates FAIIR’s final performance after refinement based on incorporating the experts' blind responses and adjusting the classification decision boundary to be less biased towards predicting the most common tags. Comparing the original threshold "FAIIR vs. Experts" and the updated threshold (UT) "FAIIR (UT) vs. Experts" in Figure \ref{table:survey_b_consensus}, average precision (0.66 $\pm$ 0.2) is improved while average recall (0.76 $\pm$ 0.14) is decreased. F1-score also improves for the full agreement on primary issue tags (0.53) and partial agreement via majority vote (0.6) settings, demonstrating benefits in the strictest consensus measures.

  \vspace{6mm}

\subsection{FAIIR Performance in Silent Trial Consistent with Development Phase}
Conversational data for the silent testing phase comprised 84,832 conversations collected between February and September 2023. The distribution of issue tags can be seen in Figure \ref{figure:silenttestingdistribution} in Appendix \ref{appendix::silenttestingtable}, demonstrating that the silent testing dataset remains largely consistent with the data used for developing the FAIIR tool. However, certain tags like \textit{DNE} are more common in the former dataset. In addition, this data naturally includes more up-to-date topics and events of 2023 such as natural disasters and political crises. Table \ref{table:mainresultslongformer} presents the evaluation performance of FAIIR predictions for each issue tag and Figure \ref{table:LongformerOverallResults} (right) indicates averaged metrics across all issue tags for three thresholds: 0.25, 0.5, and the adjusted threshold based on expert evaluation and assessment. An expected slight decrease in performance is noted and is attributable to the differences in distributions between the development and silent testing datasets. Similar to the retrospective testing setting, the majority of the AUC ROC scores are above 0.9, with the lowest score being 0.73. The sample averaged precision, recall, and F1 scores are 0.57, 0.79, and 0.62, respectively, for a threshold of 0.25, compared to 0.58, 0.81, and 0.64 for the retrospective values, indicating a drop of less than 2\%.

\section{Discussion}\label{sec12}
We have successfully demonstrated the viability of employing an NLP-based frontline assistant tool to augment CRs by identifying the issues that service users may be experiencing in text-based support conversations. After analyzing a textual conversation, the FAIIR tool is able to recommend potential issues from a list of 19 predefined tags. FAIIR achieved an accuracy of 94\%, a sample average F1-score of 64\%, and a sample average recall score of 81\% on the development set. This is a strong performance given the inherent subjectivity and noisiness of the data. Importantly, in the context of our fairness analyses, the FAIIR tool demonstrated equitable performance across all demographic groups of service users. 

Our study demonstrates the robustness and generalizability of the FAIIR tool, which is built upon recent advances in LLMs. In our silent testing phase, we observed less than a 2\% drop in sample average precision, recall and F1-score, demonstrating the tool's strong potential for real-time deployment.

Our investigation revealed that domain adaptation through self-supervised learning significantly enhances tool performance, especially in supervised tasks and when addressing label biases. This finding is relevant to our study, where we noted potential biases in the original labelling process, as each conversation was labelled by a single annotator without subsequent review. Consequently, a large annotated dataset for a supervised task with multiple annotators per conversation would be optimal. However, given the scale of our study with over 780,000 individual conversations, extensive manual annotation was not feasible. We sought expert assessment and evaluation for edge-case conversations to enrich our ground truth annotations and explore the benefits of contextual learning. The experts' responses showed an overall agreement of 90.9\% with the FAIIR tool's predictions. Notably, expert agreement with FAIIR exceeded their agreement with the original labels. This observation can be attributed to the extensive self-supervised training of FAIIR, which equipped the model with a strong contextual understanding. This training allowed the model to grasp the nuances and relationships between words and phrases within the context. The expert evaluation not only demonstrated the strong performance of the model in context but also provided a valuable source of ground-truth issue tags for further model refinement. After refining the FAIIR tool by leveraging the blind survey outcomes, the precision metric exhibited a 4\% increase.

The distribution of issue tags in our dataset was highly imbalanced, with the most frequent tag, \textit{Anxiety/Stress}, appearing in more than 244,000 conversations, while the least frequent tag, \textit{Prank}, appears in only 2,800 conversations. This imbalance posed significant challenges during our model's training, which explains the relatively poorer performance of FAIIR across less-represented issue tags. For instance, the model encountered difficulties with applying tags such as \textit{Other}, \textit{Abuse}, \textit{Emotional}; \textit{Abuse, Physical}; \textit{Isolated}, \textit{Prank}, and \textit{Testing}. Some of these challenges also stem from the inherent vagueness of certain tags. For example, \textit{Isolated} can apply to a broad spectrum of conversations, but its relevance may be selectively applied. The \textit{Other} tag presents similar difficulties, encompassing anything outside the scope of the set 19 issue tag categories. Tags like \textit{Prank} and \textit{Abuse, Physical} also suffered in performance due to their rarity in the dataset, making it challenging for the model to adequately learn to recognize these instances. While we implemented imbalanced learning techniques such as re-weighting and balanced samples, further techniques may be necessary to enhance performance. However, despite these challenges, the tool demonstrated exceptional performance, above that of random guessing.

Another challenge we encountered was the diversity and extensive length across conversations in our dataset. This was largely due to the therapeutic nature of the discussions, which primarily focused on mental health support and recovery. To manage this diversity, we established a maximum token length of 2,000, covering approximately 95\% of all conversations. This threshold not only accommodated the majority of our conversations, but also optimized the batch size for more efficient training, thereby accelerating the training process. The decision to use a Longformer model was primarily influenced by its well-suitedness for our context: its design for longer sequences and our need to handle lengthy sequences effectively made this model the correct choice. 

An additional obstacle we faced was the varying quality of conversations, including differences in language use, grammar, and the presence of noise such as typos or slang. Addressing these issues required extensive pre-processing efforts, which introduced subjectivity and potential bias into the data. 

In conjunction with the identification of primary issue tags, we established an explainability pipeline to facilitate the extraction of contextual keywords, referred to as "natural keywords" from each conversation. These keywords are dynamic tokens associated with the specific main issue tags in a conversation. To streamline the processing of natural keywords and facilitate communication with subject matter experts, we conducted word embedding visualization and bi-gram analysis at the aggregated level to demonstrate semantic relationships and word proximity in a specific context (supplementary materials). However, it is crucial to note that the reliability and meaningfulness of these natural keywords require further rigorous assessment, of which our future work will be comprised.

Our study's primary limitation is its reliance on a predefined set of 19 issue tags for the identification of topics within a conversation. This limitation restricts the model's ability to extract information beyond the predefined list. In conversations of this nature, CRs have identified their eagerness to delve into dynamic and natural issue tags that are more youth-centred, as opposed to being limited by this predefined list. Such a dynamic model could permit for change over time or based on the context of the conversation, allowing for the tool to adapt to new issues it may not have encountered during training.

\subsection{Future Directions}\label{futuredirections}
Our immediate next step involves deploying the FAIIR tool for real-time issue tag identification. Simultaneously, we aim to enhance the pipeline by incorporating generative language models and decoder-based models to further improve dynamic issue tag predictions. Additionally, we hope to evaluate the usability and validity of the identified natural keywords via a panel of subject matter experts. 

Given its positive performance and adaptiveness, the FAIIR tool demonstrates promise in application to not only the use case of issue classification described at length in this paper, but also more broadly within the mental health support context. Directly relevant applications of the FAIIR tool may include streamlining triage processes for crisis lines by identifying topics at hand, and increased robustness in flagging service users at risk. Via future works, we aim to explore the utility of the FAIIR tool in varied support contexts, building towards our goal of supporting both users and providers in leveraging NLP tools to their benefit.

\section{Conclusion }\label{conclusion}
In conclusion, the rising demand for youth mental healthcare and crisis support has become a pressing concern for both healthcare providers and users. This growing need has prompted active efforts to develop and deploy safe, trustworthy, and transparent conversational AI solutions that support providers by reducing the administrative burden inherent to providing support and guidance to young individuals facing mental health challenges. Our study contributes to the ongoing exploration of solutions by showcasing the development and evaluation of a front-line conversational agent assistant tool, while sharing lessons learned with the broader community. By demonstrating the effectiveness and feasibility of such solutions, our study paves the way for broader adoption and implementation of conversational AI models in mental health and crisis support services.

\section{Online Methods}\label{sec11}

\subsection{Study Design and Setting}
The study comprised two phases. The first phase involved building FAIIR for the identification of issue(s) that a young person might be experiencing from their textual conversations with trained CRs. Identification was performed using a list of 19 predefined issue tags. This phase laid the foundation for our research, where we diligently developed, fine-tuned, and evaluated FAIIR's capacity as an NLP tool to understand and predict issues. In the second phase, we validated the model's efficacy and accuracy through testing with domain experts and silent testing. This phase confirmed the practical applicability and real-world utility of FAIIR for CRs. Both phases utilized conversations related to crisis support services at KHP. 

\subsection{Curation of Study Dataset}
The primary conversational dataset used for building and evaluating the NLP models of FAIIR was comprised of 703,975 unique, scrubbed, multi-turn dialogue instances between service users and CRs via SMS from January 2018 until February 2023. An additional batch of 84,832 conversations from February to September 2023 was used for silent testing. It is important to note that some of these dialogues may originate from service users who engage in multiple instances of interaction with CRs; however, encounters from the same individual are not linked. In total, the training data represented conversations with 340,512 individual service users and 7,937 CRs. The silent testing data represented 57,031 unique service users and 2,038 CRs, with expected overlap between the individuals represented in both datasets. 

At the end of each conversation, service users are asked to fill out an optional demographic survey. The demographic survey captures information including the helpfulness of the conversation to the user and demographics including their age range, ethnicity or cultural group, identification with any of 10 identity groups (e.g. newcomer, refugee, deaf, blind, people with disabilities), and setting of current living (i.e. city, rural area, First Nations Reserve). Approximately 17\% of service users typically complete this survey, most of whom identify as female, heterosexual and of European ancestry. Most conversations are flagged as medium-risk as shown in Figure 1 with the distribution of priority labels across the main conversational dataset, according to the priority flagging methods further described in Appendix \ref{secA3}.

A total of 19 pre-defined issue tags currently serve to describe the range of topics raised by a user during a conversation, including topics such as \textit{Depressed}, \textit{Anxiety/Stress}, and \textit{Gender/Sexual Identity}. Upon conversation conclusion, a CR manually assigns at least one available issue tag(s) to the conversation. Metrics related to tags are used in aggregate for insight generation, to best follow trends in youth issues, support CRs, as well as reporting to funders and other agencies. It is important to note that this labelling process is carried out by CRs at their own discretion, and according to their training. Due to limited resources and large volumes of service user inquiries, issue tags typically do not undergo additional review.

Data was anonymized, undergoing a process of scrubbing identifying information such as names and locations, which were automatically replaced with the placeholder \textit{[scrubbed]}. In many instances, complete phrases and sentences were scrubbed due to the anonymization process. This process therefore introduced some noise due to the unintentional removal of harmless words, like \textit{turkey}.

\subsection{Development of the FAIIR Tool}
We framed the issue tag identification task as a multi-label classification problem, where multiple labels can be assigned to a single instance. The development of the classifier followed two distinct stages. In the first stage, we compared and evaluated various pre-trained transformer-based language models, fine-tuning them on a randomly selected subset of the data for classification. Pre-training involves training models on large-scale text corpora to learn general language patterns, while fine-tuning adapts these models to a specific task using a smaller, task-specific dataset. The second step involved domain adaptation, refining the model to better capture the nuances of youth mental health conversations. This was achieved through additional pre-training and fine-tuning on the full baseline training dataset, ensuring the classifier effectively recognized context-specific language patterns and issues relevant to the domain.\\
\newline
\textbf{Step 1: Model Comparisons}\\
We explored four primary variants of transformer models for processing lengthy documents and task-oriented conversational data. These models fall into two categories of ``encoder-only'' models - designed primarily for classification tasks and ``encoder-decode'' models - which process input text using an encoder and generate output using a decoder \citep{NIPS2017_3f5ee243}. The models evaluated included Longformer \citep{Beltagy2020LongformerTL}, an encoder-only model with 149M parameters, Conversational BERT, an encoder-only model with 110\textit{M} parameters \citep{convBert, devlin2018bert}, DialogLED, an encoder-decoder model with 139\textit{M} parameters \citep{Zhong2021DialogLMPM}, and MVP (Multi-task superVised Pre-training), an encoder-decoder model with 406\textit{M} parameters \citep{Tang2022MVPMS}. During fine-tuning, these models, we incorporated a classification head, a single-layer neural network that converts the model's output into probability scores for each class label. In encoder-only models, this layer was applied to the [CLS] token, which represents the entire input. For encoder-decoder models, classification was based on the [EOS] token (DialogLED) or the first token in the sequence (MVP), following established conventions \citep{kementchedjhieva-chalkidis-2023-exploration}.

All four models were fine-tuned on 50,000 conversations randomly sampled from the dataset. We built a 60/20/20 stratified training/validation/test. The fine tuning on the full dataset took approximately 12 hours per epoch on four A10 NVIDIA GPUs (24GB VRAM) with 16 CPU cores, each with an effective batch size of 16. Learning rates were tuned within the range of $1e^{-5}$ to $3e^{-5}$. Max token lengths were applied for BERT, DialogLED, and MVP, while Longformer was capped at 2,048 tokens for efficiency. This limits the length of input text that can be provided but provides faster processing. The optimal training durations for each model were determined through basic hyperparameter tuning (used to find optimal parameters such as learning rate and batch size), resulting in two epochs (training cycles) for BERT, three epochs for DialogLED, five epochs for Longformer, and two epochs for MVP. Threshold selection was a key consideration in determining how labels were assigned. Since this is a multi-label classification task, where each conversation can have multiple assigned tags, we experimented with different threshold values to optimize the balance between precision and recall. We systematically evaluated thresholds ranging from 0.25 to 0.5 on a validation set, measuring their impact on classification performance. Our analysis indicated that a threshold of 0.25 yielded the best trade-off between precision and recall, particularly for underrepresented labels.\\ 
\newline
\textbf{Step 2: Final Model Development and Optimization}\\
For our final model, we employed an ensemble approach combining three Longformer models, each with slightly different initialization and fine-tuning settings. The choice of Longformer as our primary model was based on its superior performance and its capacity to effectively capture long conversations. Each Longformer underwent initial pre-training using the same approach, which included masked language modelling, where a portion of words in each conversation was masked, and the model learned to predict them, on the full baseline training dataset. We applied masking to 15\% of tokens per conversation and pre-trained the models for one epoch with a maximum sequence length of 1,500 tokens. AdamW \cite{loshchilov2018decoupled} was used as the optimizer, and a linear scheduler with 500 warm-up steps was applied to improve training stability. Gradient accumulation was used to maintain an effective batch size of 64, ensuring efficient use of GPU resources. This pre-training step required approximately 24 hours.

Following the pre-training task, the Longformer models were fine-tuned on a label-balanced training/validation/test data split (60/20/20). Per recommendation of our domain experts, we incorporated additional context information related to the conversation's priority. Therefore, the beginning sentence included the statement: "This conversation is of $<<X>>$ priority" with $X$ representing one of the three priority levels assigned to each conversation. More on this process of generating these levels is discussed in Appendix \ref{secA3}. Each Longformer model was fine-tuned for a maximum of three epochs using a batch size of 16, managed through gradient accumulation, with a learning rate set to $2e^{-5}$. Standard Binary Cross Entropy loss was applied during the fine-tuning, with oversampling of conversations with less common issue tags specifically implemented on two of the ensemble models to address the class imbalance. We used AdamW as the weight optimizer and implemented a linear scheduler with the initial 20\% of training steps.

\subsection{Evaluation of FAIIR Predictions}
Upon completion of the development of the FAIIR tool, we conducted two independent experiments to evaluate its efficacy and performance in generalization. The experiments included both expert assessment and silent testing of the tool and its predictions. \\

\noindent \textbf{Expert Assessment and Evaluation}\\
Methods for expert assessment for FAIIR included conducting an evaluation survey completed by CRs. We invited 12 trained CRs to review 40 challenging conversations. The conversation selection criteria were diverse, focusing on those with more than 4 issue tags and including ambiguous cases where FAIIR's predictions were confident but incorrect based on our ground-truth labels (see Appendix \ref{secA1} for further details about conversation selection). Our hypothesis was that for these edge cases, FAIIR requires a deep and nuanced understanding to perform well. Thus, our goal was to assess the model's ability to identify all relevant issue tags and navigate language nuances. 

Each conversation was independently reviewed by six CRs, divided into two groups. In the "open review", three CRs reviewed conversations with FAIIR’s predicted issue tags explicitly provided. This approach aimed to evaluate whether the model’s predictions were helpful, misleading, or partially correct in identifying the core issues within each conversation. CRs could either agree or disagree with the predicted tags and suggest corrections or refinements where necessary.

In the "blind review", the remaining three CRs reviewed the same conversations without any prior exposure to FAIIR’s predicted tags. Instead, they independently identified issue tags based solely on the conversation content. Furthermore, they categorized the identified tags into primary issue tags (representing the most pressing concerns) and secondary issue tags (minor but relevant concerns). This approach established a baseline for comparison against FAIIR’s predictions, ensuring that human assessments were made without any influence from the model.

The following five criteria were established to develop a consensus measure for comparison in the blind review setting, which is more challenging than the open review setting. Since human annotations categorize issue tags as primary (most pressing concerns) and secondary (minor but relevant concerns), we evaluated agreement with FAIIR’s outputs based on these distinctions. Notably, FAIIR does not explicitly differentiate between primary and secondary issue tags—all predicted tags are treated equally. Therefore, for the purpose of comparison, we assessed agreement by mapping FAIIR’s predicted tags to human annotations and measuring alignment using the following criteria:
\begin{itemize}
    \item Full agreement on primary issue tags (FA: 1\(^\circ\)) - all primary issue tags identified by human annotators are also predicted by FAIIR.
    \item Partial agreement on primary issue tags via majority vote (PA: 1\(^\circ\) Maj.) - the majority of human annotators agree on a set of primary issue tags, and these tags overlap with FAIIR’s predictions.
    \item Partial agreement on primary and secondary issue tags via majority vote (PA: 1\(^\circ\)$+$2\(^\circ\) Maj.) - the majority of human annotators agree on a set of both primary and secondary issue tags, and these overlap with FAIIR’s predictions.
    \item Full agreement on primary issue tags via at least one vote (FA: 1\(^\circ\)\(\geq\) 1) - at least one human annotator identified a primary issue tag that is also predicted by FAIIR.
    \item Full agreement on primary and secondary issue tags via at least one vote (FA: 1\(^\circ\)$+$2\(^\circ\)\(\geq\) 1) - at least one human annotator identified a primary or secondary issue tag that is also predicted by FAIIR.
\end{itemize}

\noindent
\textbf{Model Refinement - Modifying the decision boundary}\\
In our evaluation experiments, model refinement involved adjusting the decision boundary (threshold cutoff) to strike a balance between recall and precision. In most experiments, the FAIIR tool's predictions showed lower precision compared to recall, so we adjusted the threshold to reduce its frequency of outputting the most common tags while lowering the threshold for rare tags. For example, we set the threshold to 0.4 for the three most frequent classes: \textit{Anxiety/Stress}, \textit{Depressed}, and \textit{Relationship}. For the next two most frequent classes, \textit{Suicide} and \textit{Isolated}, we adjusted the threshold to 0.3. These five classes encompass the majority of predicted issue tags from the model, hence we targeted them for increased thresholds. The remaining tags were set at a lower threshold of 0.2 to enhance the model's ability to capture them effectively.\\
\newline
\textbf{Silent Testing}\\
We conducted silent testing to assess the FAIIR tool's generalization performance on new and recent batches of conversations received over 8 months. Testing on 84,832 conversations that occurred between February to September 2023 served as a valuable representation of how the model adapts to and handles the ever-changing landscape of real-world dialogue. In addition to model evaluation, we implemented refinements as explained earlier to strike a balance between precision and recall.

\subsection{Outcome Interpretation and Visualization}
We leveraged layer-integrated gradients \citep{sundararajan2017axiomatic}, a technique to understand which tokens in any given conversation hold the utmost relevance to the predicted issue tags. By doing so, not only can we identify the most pertinent words associated with the primary conversation topics but we also gain insight into the model's decision-making process, thereby enhancing its overall explainability. To achieve this, we computed an attribution score, which are analogous to an importance score, for each token in the conversations using the final model and tokenizer with respect to the sets of predefined issue tags. Attributed tokens along the axis of issue tags that surpass a predefined threshold are selected as the most relevant words. We refer to these as "natural keywords". The term 'natural' is used because these keywords are not predefined; they are dynamic and align with the nuances of natural language. 

To create a cleaner set of natural keywords from the initial set obtained using integrated gradients, we run a series of filters to remove words and symbols that are irrelevant or do not add meaning or additional insights. We automatically filtered stop words, punctuation, and special tokens for this reason. In addition, we devised a predefined word list that contains natural keywords. These keywords, while not categorized as stop words, consistently occurred very frequently across virtually every issue tag (e.g., User, Hello, Connect). Any keywords that fall within this list were also filtered out. Finally, we filter keywords according to their part of speech tags, removing any that fall within a defined set of categories. We filtered conjunctions, determiners, prepositions, modal auxiliary words, and the majority of verbs along with many other categories, yielding keywords that are primarily nouns and adjectives.
 
To streamline the processing of natural keywords and facilitate communication with knowledge experts, we conducted word embedding visualization and bi-gram analysis at the aggregated level to demonstrate semantic relationships and word proximity in a specific context.


\noindent  \section*{Declarations}\label{sec5}

\noindent \subsection{Ethics Statement}\label{sec5.1}
Kids Help Phone (KHP) is deeply committed to the ethical and responsible use of data to enhance our services for youth, recognizing the importance of ethical principles in maintaining the trust of those we serve, especially the most vulnerable. This paper is aimed exclusively at applied research to improve service delivery and accessibility, with a special focus on the ethical application of Artificial Intelligence (AI) to benefit our service network and frontline staff. Through this collaboration, we are dedicated to developing technological tools that provide a personalized and user-friendly experience for those seeking help. Upholding the privacy and confidentiality of our service users is paramount; we adhere to an ethical statement aligned with KHP’s privacy policy (https://kidshelpphone.ca/privacy-policy/), including consent notice for research and rigorous data minimization. Our processes are transparent and accountable, compliant with Canadian privacy regulations. We meticulously remove all direct identifiers from research data, adhering to industry standards for data anonymization, and securely store all research data within KHP’s infrastructure. This reflects our commitment to the highest standards of data security, confidentiality, and ethical practice. By prioritizing ethical data use, KHP can leverage research to improve our services and deliver the best possible support for youth across Canada, embodying our commitment to integrity, respect, and responsibility in every action we take.

\subsection{Acknowledgements }
We acknowledge the critical support and engagement of Kids Help Phone's team of texting staff and crisis responders, who underpinned, guided and informed this work. We also wish to express gratitude to Kids Help Phone's Innovation and Data team, as well as their Brand, Storytelling and Communications team, who supported in editing.

\subsection{\noindent Contributions\label{sec5.2}}
SO, MN, JR, JY and ED drafted the main manuscript. SO and ED prepared the figures and methods. AL, JR and DM provided subject matter expertise. AL, XZ, DM, DP, RS and ED provided supervision and guidance. All authors reviewed the manuscript.

\subsection{\noindent Competing Interests}
The authors declare no competing interests.

\noindent \subsection{Data and Code Availability}
The data and code that supports the FAIIR tool and this study overall are not openly available due to reasons of sensitivity. These data and code are located in controlled access storage at Kids Help Phone. Full code is available at the private GitHub https://github.com/KidsHelpPhone/AI-ML/tree/main/FAIIR\%20V1\%E2\%80\%9C upon request. Please contact \href{mailto:contact@kidshelpphone.ca}{contact@kidshelpphone.ca} for more information.

\begin{appendices}

\section{Issue Tag Definitions}\label{secA00}

\noindent\begin{minipage}{\linewidth}
\centering
\captionof{table}{Definitions of each of the 19 issue tags.}\label{table:IssueTagDefs}\resizebox{\linewidth}{!}{%
\begin{tabular}{l|l} 
\hline
\textbf{Issue Tags} & \textbf{Definition} \\ 
\hline
\textit{3rd Party} & Seeking support for another person who is in crisis (the third party). \\ 
\hline
\textit{Abuse, emotional} & \begin{tabular}[c]{@{}l@{}}Emotional abuse, also known as psychological abuse, is when someone threatens, bullies and intimidates another person; Can include things like name calling/insulting someone, \\humiliating a person in public, ridiculing someone, abusing someone else in front of a young person, threatening to leave someone,\\threatening to hurt someone/somtething a person cares about, telling someone they're a bad person, etc.\end{tabular} \\ 
\hline
\textit{Abuse, physical} & \begin{tabular}[c]{@{}l@{}}Physical abuse includes physically harming a person in some way\\~(e.g., slapping, biting, scratching, biting, strangling, throwing objects, hitting, punching, shaking, choking, using an object to cause pain/harm)\end{tabular} \\ 
\hline
\textit{Abuse, sexual} & \begin{tabular}[c]{@{}l@{}}Sexual abuse is when a person uses sexual acts as a way to demonstrate power or authority over someone else or when and adult is involved in any kind of sexual activity with a minor.\\~Sexual abuse often involves physical contact, but it can also happen without touching; This can include touching a person's genitals or making them touch someone else's genitals, \\having sex or trying to have sex with a young person, taking off a person's clothes or forcing them to watch as someone else takes off their cloths (especially in a sexual way), \\making sexual comments about a person's body, forcing someone to watch others have sex, making someone watch pornography, taking pictures of someone without their clothes on, etc.\end{tabular} \\ 
\hline
\textit{Anxiety/Stress} & Feelings of worry, nervousness and panic; A reaction to a
perceived threat; A perception of pressure and/or a physical response in body to pressure/heightened emotions \\ 
\hline
\textit{Bully} & \begin{tabular}[c]{@{}l@{}}When someone intentionally uses their power to hurt, frighten, exclude or insult someone in-person; \\Harmful and unwanted behaviour (e.g., can be physical, emotional, social, internet-based) that can make someone feel embarrassed, offended, intimidated or unsafe\end{tabular} \\ 
\hline
\textit{Depressed} & \begin{tabular}[c]{@{}l@{}}Depression here is not meant to represent a formal diagnosis, about feeling depressed/sad/etc.;\\~Persistent feelings of sadness: may include inactivity, hopelessness; losing interest in things you enjoy; crying a lot; feeling helpless or hopeless; feeling sad, down or low\end{tabular} \\ 
\hline
\textit{Did Not Engage (DNE)} & Did not respond to greeting or first message. \\ 
\hline
\textit{Eating Body Image} & \begin{tabular}[c]{@{}l@{}}Body image is how someone perceives their body/how they look, can include discussing things like body dysmorphia, how their body looks in different clothing/fashion; \\May be positive or negative perceptions .A range of irregular eating behaviors that may or may not warrant a diagnosis of a specific eating disorder; \\May include restricting eating, excessive dieting, purging, binging, having changes in appetite (e.g. eating less, eating more, etc.)\end{tabular} \\ 
\hline
\textit{Gender/Sexual Identity} & \begin{tabular}[c]{@{}l@{}}Sexual orientation, gender identity, gender expression, gender transition, or other topic relating to gender and/or sexual identity/orientation;\\~A person’s gender identity may be the same as or different from their assigned sex at birth; Could be about someone's own identity or questions about it in general;\end{tabular} \\ 
\hline
\textit{Grief} & Mourning or grieving a loss of any kind (e.g., a person, a significant change, an animal, an experience) \\ 
\hline
\textit{Isolated} & Feeling alone, companionless, unsupported, isolated; withdrawing from friends and family \\ 
\hline
\textit{Other} & Issue does not fit any of these categories. \\ 
\hline
\textit{Prank} & Playing a practical joke. Not in crisis/seeking support. \\ 
\hline
\textit{Relationship} & \begin{tabular}[c]{@{}l@{}}Relationships - With family member(s) (e.g., parents/guardians, siblings, etc.)\\ e.g., parents/guardians, siblings, etc.; Concerns, dynamics, relationship building or breakdown, stress, or preoccupation with family;\\~for example, talking about healthy/unhealthy relationships with specific people or in general, talking about toxic relationships, etc.\\ Relationships - With friend(s)/peer(s)\\ e.g., classmates, colleagues, team mates, community friends, etc.; Concerns, dynamics, relationship building or breakdown, stress, or preoccupation with friends/peers;\\~for example, talking about healthy/unhealthy relationships with specific people or in general, talking about toxic relationships, etc.\\ Relationships - With partner(s) (e.g., boyfriend/girlfriend, spouse, romantic partners)\\ e.g., boyfriend/girlfriend, spouse, romantic partners; Concerns, dynamics, relationship building or breakdown, stress, or preoccupation with romantic relationships; \\for example, talking about healthy/unhealthy relationships with specific people or in general, talking about toxic relationships, etc.\\ Relationships - Other relationship(s)\\ e.g., strangers, online relationships, professional relationships, etc.; Concerns, dynamics, relationship building or breakdown, stress, or preoccupation with other relationships;\\~for example, talking about healthy/unhealthy relationships with specific people or in general, talking about toxic relationships, etc.\end{tabular} \\ 
\hline
\textit{Self Harm} & \begin{tabular}[c]{@{}l@{}}When a person purposely hurts their body without trying to kill themself (Non-fatal self-injury).\\~People who self-injure might cut, burn, hit or bite themselves, pull out their own hair or pick at sores on their skin.\\~Many people who self-injure say hurting themselves gives them an immediate sense of relief.\end{tabular} \\ 
\hline
\textit{Substance Abuse} & Addiction to legal/illegal substances~ e.g., Alcohol, prescription medication \\ 
\hline
\textit{Suicide} & Intentionally and voluntarily attempting, self-directed injurious behavior with an intent to die OR suicidal ideation-thinking about, considering, or planning suicide. \\ 
\hline
\textit{Testing} & Testing the system (e.g., asking if this number is real, KHP employee testing functionality of the service) \\
\hline
\end{tabular}
}
\end{minipage}

\section{Process of Selecting Model Architecture}\label{secA0}

The utilization of transformer-based models, such as Longformer, for classifying clinical or conversational data has been extensively explored in the literature. Studies, such as \citet{Li2022ClinicalLongformerAC}, have consistently demonstrated that Longformer models outperform shorter-sequence transformers like ClinicalBERT \cite{clinicalbert} in various downstream tasks, including clinical document classification. Similarly, in another study by \citet{dai-etal-2022-revisiting}, which evaluated different approaches for classifying long documents using transformer architectures such as Longformer, it was concluded that employing transformer-based models designed for longer sequences is more effective and efficient than using shorter-sequence models like BERT. This finding is particularly relevant as BERT is constrained by a 512-token limit which prevents the processing of any text that has a token length greater than the limit, while Longformer's capability to handle longer sequences (up to eight times longer) proves advantageous for tasks requiring a broader context, such as analyzing lengthy conversational data. Other work such as those of \citet{wang-etal-2021-cs}, \citet{Zhong2021DialogLMPM} and \citet{ji2023domainspecific}, highlight the significance of additional pre-training methods, such as masked language modelling and next-turn prediction, especially in the context of dialogue data. Both studies emphasize the differences between general domain language and dialogue, indicating that pre-training on a large corpus of domain-specific dialogues can significantly improve performance on downstream dialogue tasks. Particularly \citet{Zhong2021DialogLMPM} demonstrates the benefits of pre-training Longformer using dialogue-specific window-based denoising on lengthy dialogues, resulting in a substantial improvement in state-of-the-art tasks such as long dialogue understanding. Lastly, \citet{ji2023domainspecific} pre-train RoBERTa \citep{liu2019roberta}, Longformer and XLNet \citep{10.5555/3454287.3454804} on mental healthcare domain data for the task of mental health classification, achieving superior results in most cases to the base models, demonstrating the effectiveness of significant pre-training on mental healthcare domain data in related downstream tasks.

Table \ref{table:preliminary_model_selection} compares the performance of the four preliminary models, including Longformer, Conversational BERT, DialogLED, and MVP, all chosen for their suitability for handling either conversational data or long-documents. We used five metrics to evaluate model performance on the test data. The first metric is the standard classification “accuracy” which considers the total percentage of all of the 19 tags predicted by the model correctly for each instance across the full dataset. In this way, to attain full accuracy for a given conversation, the model must predict all of the correct set of tags assigned to the conversation, and not mistakenly predict any tags that are not in the correct set of tags.
Due to the sparsity of assigned tags, with conversations tending to be tagged with only a few tags out of the 19 total, a classifier can attain a high accuracy by not predicting any tags, hence we use a second metric which is referred to as “exact accuracy”, and assesses correctness based on the percentage of conversations where all the predicted issue tags are correct. As such, a single misidentified tag for a conversations means it is classified as an incorrect prediction. In addition to accuracy, we use three other metrics which we call “sample average precision”, “sample average recall”, and “sample average F1-score”. 

In the context of multi-label classification, the sample/example-based average calculates the three scores for each sample then averages the scores across all samples. For each sample, the entire set of predicted tags is considered in the calculation of the three scores without isolating each individual tag type, and are compared with the full set of true labels. This is unlike micro-averaging where the scores are calculated globally across all of the total true positives, false negatives and false positives, or macro-averaging where the scores are calculated across all of the true positive, false positive, and false negative counts for a specific tag $i$ before taking the unweighted mean across the scores for all tags. 
This method of averaging provides a representative result for the entire distribution after assessing scores for each sample individually.The metrics displayed in Table \ref{table:preliminary_model_selection} are averages across all issue samples. 

Both Longformer and Conversational BERT exhibit comparable high performance (Accuracy: 0.94 and sample average F1-score: 0.56). Conversational BERT offers the advantage of being pre-trained on an extensive corpus of conversation data, while Longformer excels in capturing longer sequences. Therefore, we leveraged Longformer due to the nature of our conversations (long sequences), with the intention of performing domain adaptation akin to Conversational BERT to improve its performance. The remaining two models, based on encoder-decoder architectures, underperformed (sample average F1-score $<$0.35) primarily because they were not originally designed for this particular multi-label classification task. Furthermore, both encoder-decoder models encountered significant practicality issues related to exceptionally lengthy training times and resource limitations, necessitating use of small batch sizes and long inference times. As a result, they were deemed sub-optimal choices for this specific task.

\begin{table}
\centering
\caption{Summary of the performance of four transformer models fine-tuned for multi-label classification tasks on 50,000 conversations.}\label{table:preliminary_model_selection}
\begin{tabular}{lllll} 
\hline
                               & \textbf{Longformer} & \textbf{BERT} & \textbf{DialogLED} & \textbf{MVP}  \\ 
\hline
\textbf{Accuracy}              & 0.938               & 0.936         & 0.922              & 0.351         \\
\textbf{Exact Accuracy}        & 0.336               & 0.339         & 0.206              & 0.000         \\
\textbf{Sample Avg. Precision} & 0.660               & 0.650         & 0.430              & 0.200         \\
\textbf{Sample Avg. Recall}    & 0.530               & 0.540         & 0.320              & 0.820         \\
\textbf{Sample Avg. F1}        & 0.560               & 0.560         & 0.350              & 0.310         \\
\hline
\end{tabular}
\end{table}

\section{Survey Conversation Selection Details}\label{secA1}

20 of the 40 total conversations annotated were picked at random from the test set to get a representative sample of the data. All 20 conversations were originally labelled with 4 or more different issue tags. The reason for selecting these conversations is their coverage of a wide range of issues: the potential for annotators picking a different set of issue tags from each other is high, and perspectives to determine which issue tags truly apply to the conversation were of upmost importance. This was also important in evaluating whether the model was able to grasp all nuanced issue tags that may apply less directly to a given conversation. 

The remaining 20 of the 40 conversations were mostly originally tagged with three or less issue tags, in an effort to promote a balance between conversations with many tags and those where only a small number may apply.

Of these 20, handfuls of conversations were selected according to several differing criteria. A number were selected manually to cover all of the 19 different issue tags, in an effort to build consensus in the identification of all tags for the model to reference. A small sample of conversations were also selected as purposefully ambiguous cases: mainly long conversations which were only annotated with 1 or 2 issue tag(s). Although the issue tag(s) assigned at baseline were typically correct, these conversations were an opportunity to gain consensus on a spectrum of more nuanced tags for the purposes of model fine-tuning. The last few conversations were handpicked because they were perceived to be mislabelled in some way. For these conversations, the original issue tag(s) assigned appeared incorrect or incomplete, in that there was another key issue tag missing. Consensus building is important for these examples, in order to improve the original tag(s) assigned, where incorrect. These can also be complex cases for the model to navigate, and thus served as a helpful way to evaluate the tool's performance.

\section{Priority Flag Pre-Processing}\label{secA3}

 At the start of each conversation, the system generates a priority flag based on the service user's first few words. Service users are then triaged into categories of either high, medium, low-risk, or “no ground truth” via an algorithm owned by Crisis Text Line. Medium risk is assigned when a user expresses suicidal thoughts or self harm, and high risk is assigned when an individual is deemed to be an “imminent risk”, defined as having a combination of suicidal thoughts, a plan, access to means, and a $0-48$ hour timeline to end their life. The presence of any 56 English or 73 French words in an initial message from a user leads to their automatic triage to a higher priority level. According to the distribution in Figure 1, the vast majority of conversations (87\%) were flagged as medium-risk, with about 13\% being flagged as high risk. Almost no conversations (0.0001\%) were flagged as low-risk. 

To assess how the FAIIR model performs across the different assigned priority levels, similar to Table 1, we collected fine-grained performance of the main FAIIR tool across our main metrics on the retrospective test set (n=140,795). We divided the conversations in the test set into two main priority levels ("Medium" and "High") and report the results across the two thresholds (0.25 and 0.5) in Table \ref{table:priorityesultslongformer}. We observe that performance across the two risk categories has little variation, meaning there is little bias towards conversations according to priority and that the model is able to handle these different levels accordingly.

\begin{table*}
\centering
\caption{The fine-grained performance of the main FAIIR tool (ensemble of Longformer models) on the retrospective test set (n=140,795) across the conversations divided into two main priority levels in the dataset ("Medium" and "High"). Results are provided as averages across all samples for classification thresholds of 0.5 and 0.25. The Weighted Avg. column provides the sample-averaged results for each respective metric.}
\label{table:priorityesultslongformer}
\resizebox{\linewidth}{!}{%
\begin{tblr}{
  row{2} = {c},
  row{10} = {c},
  column{4} = {c},
  column{5} = {c},
  column{6} = {c},
  column{7} = {c},
  column{8} = {c},
  column{9} = {c},
  column{10} = {c},
  column{11} = {c},
  column{12} = {c},
  column{13} = {c},
  column{14} = {c},
  column{15} = {c},
  column{16} = {c},
  column{17} = {c},
  column{18} = {c},
  column{19} = {c},
  column{20} = {c},
  column{21} = {c},
  cell{1}{1} = {c=2}{},
  cell{2}{1} = {c=22}{},
  cell{3}{3} = {c},
  cell{4}{1} = {r=3}{},
  cell{4}{3} = {c},
  cell{5}{3} = {c},
  cell{6}{3} = {c},
  cell{7}{1} = {r=3}{},
  cell{7}{3} = {c},
  cell{8}{3} = {c},
  cell{9}{3} = {c},
  cell{10}{1} = {c=22}{},
  cell{11}{3} = {c},
  cell{12}{1} = {r=3}{},
  cell{12}{3} = {c},
  cell{13}{3} = {c},
  cell{14}{3} = {c},
  cell{15}{1} = {r=3}{},
  cell{15}{3} = {c},
  cell{16}{3} = {c},
  cell{17}{3} = {c},
  hline{1,18} = {-}{0.08em},
  hline{2,4,7,12,15} = {1-21}{0.03em},
  hline{2,4,7,12,15} = {22}{},
  hline{3,11} = {-}{0.05em},
  hline{10} = {-}{},
}
                            &           & \rotatebox{90}{3rd Party} & \rotatebox{90}{Abuse, Emotional} & \rotatebox{90}{Abuse, Physical} & \rotatebox{90}{Abuse, Sexual} & \rotatebox{90}{Anxiety/Stress} & \rotatebox{90}{Bully} & \rotatebox{90}{Depressed} & \rotatebox{90}{DNE} & \rotatebox{90}{Eating Body Image} & \rotatebox{90}{Gender/Sexual Identity} & \rotatebox{90}{Grief} & \rotatebox{90}{Isolated} & \rotatebox{90}{Other} & \rotatebox{90}{Prank} & \rotatebox{90}{Relationship} & \rotatebox{90}{Self Harm} & \rotatebox{90}{Substance Abuse} & \rotatebox{90}{Suicide} & \rotatebox{90}{Testing} & \rotatebox{90}{Weighted Avg.} \\
\textbf{ Medium Priority }  &           &                           &                                    &                                 &                               &                                &                       &                           &                     &                                   &                                        &                       &                          &                       &                       &                              &                           &                                 &                         &                         &                               \\
                            & AUC ROC     & 0.99                      & 0.96                               & 0.98                            & 0.99                          & 0.87                           & 0.96                  & 0.96                      & 0.86                & 0.98                              & 0.99                                   & 0.95                  & 0.87                     & 0.74                  & 0.97                  & 0.90                         & 0.98                      & 0.98                            & 0.94                    & 0.97                    &       - &                       \\
Threshold - 0.25            & Precision & 0.62                      & 0.34                               & 0.32                            & 0.48                          & 0.62                           & 0.29                  & 0.68                      & 0.52                & 0.52                              & 0.54                                   & 0.34                  & 0.48                     & 0.28                  & 0.34                  & 0.65                         & 0.55                      & 0.34                            & 0.55                    & 0.42                    & 0.54                          \\
                            & Recall    & 0.94                      & 0.70                                & 0.78                            & 0.91                          & 0.82                           & 0.80                   & 0.86                      & 0.80                 & 0.89                              & 0.92                                   & 0.83                  & 0.70                      & 0.47                  & 0.67                  & 0.86                         & 0.92                      & 0.87                            & 0.84                    & 0.74                    & 0.80                           \\
                            & F1        & 0.75                      & 0.46                               & 0.45                            & 0.63                          & 0.70                            & 0.43                  & 0.76                      & 0.63                & 0.65                              & 0.68                                   & 0.48                  & 0.57                     & 0.35                  & 0.45                  & 0.74                         & 0.69                      & 0.49                            & 0.67                    & 0.53                    & 0.64                          \\
Threshold - 0.5             & Precision & 0.68                      & 0.52                               & 0.50                             & 0.58                          & 0.80                            & 0.46                  & 0.80                       & 0.69                & 0.64                              & 0.62                                   & 0.45                  & 0.66                     & 0.40                   & 0.47                  & 0.77                         & 0.63                      & 0.45                            & 0.68                    & 0.61                    & 0.69                          \\
                            & Recall    & 0.92                      & 0.50                                & 0.61                            & 0.82                          & 0.58                           & 0.67                  & 0.76                      & 0.52                & 0.81                              & 0.87                                   & 0.75                  & 0.42                     & 0.24                  & 0.56                  & 0.67                         & 0.83                      & 0.77                            & 0.65                    & 0.60                     & 0.60                           \\
                            & F1        & 0.78                      & 0.51                               & 0.55                            & 0.68                          & 0.67                           & 0.54                  & 0.78                      & 0.59                & 0.72                              & 0.72                                   & 0.57                  & 0.51                     & 0.30                   & 0.51                  & 0.72                         & 0.72                      & 0.57                            & 0.67                    & 0.61                    & 0.63                          \\
\textbf{High Priority} &           &                           &                                    &                                 &                               &                                &                       &                           &                     &                                   &                                        &                       &                          &                       &                       &                              &                           &                                 &                         &                         &                               \\
                            & AUC ROC     & 1.00                      & 0.96                               & 0.98                            & 0.98                          & 0.83                           & 0.96                  & 0.94                      & 0.80                & 0.98                              & 0.99                                   & 0.95                  & 0.84                     & 0.73                  & 0.94                  & 0.88                         & 0.95                      & 0.94                            & 0.92                    & 0.97                    &       -                          \\
Threshold - 0.25            & Precision & 0.71                      & 0.38                               & 0.37                            & 0.43                          & 0.51                           & 0.32                  & 0.51                      & 0.44                & 0.42                              & 0.43                                   & 0.34                  & 0.4                      & 0.29                  & 0.31                  & 0.53                         & 0.56                      & 0.23                            & 0.80                     & 0.36                    & 0.56                          \\
                            & Recall    & 0.98                      & 0.72                               & 0.85                            & 0.92                          & 0.68                           & 0.75                  & 0.77                      & 0.81                & 0.81                              & 0.93                                   & 0.82                  & 0.64                     & 0.30                   & 0.52                  & 0.82                         & 0.93                      & 0.81                            & 0.93                    & 0.59                    & 0.81                          \\
                            & F1        & 0.82                      & 0.49                               & 0.52                            & 0.59                          & 0.58                           & 0.45                  & 0.62                      & 0.57                & 0.55                              & 0.59                                   & 0.48                  & 0.49                     & 0.30                   & 0.39                  & 0.64                         & 0.70                       & 0.35                            & 0.86                    & 0.45                    & 0.65                          \\
Threshold - 0.5             & Precision & 0.75                      & 0.56                               & 0.53                            & 0.49                          & 0.72                           & 0.45                  & 0.65                      & 0.59                & 0.56                              & 0.48                                   & 0.45                  & 0.61                     & 0.39                  & 0.44                  & 0.68                         & 0.65                      & 0.32                            & 0.85                    & 0.56                    & 0.68                          \\
                            & Recall    & 0.96                      & 0.51                               & 0.72                            & 0.81                          & 0.37                           & 0.59                  & 0.65                      & 0.46                & 0.66                              & 0.84                                   & 0.71                  & 0.31                     & 0.10                   & 0.44                  & 0.57                         & 0.87                      & 0.64                            & 0.86                    & 0.42                    & 0.62                          \\
                            & F1        & 0.84                      & 0.53                               & 0.61                            & 0.61                          & 0.49                           & 0.51                  & 0.65                      & 0.52                & 0.61                              & 0.61                                   & 0.55                  & 0.41                     & 0.16                  & 0.44                  & 0.62                         & 0.74                      & 0.43                            & 0.86                    & 0.48                    & 0.63                          
\end{tblr}
}
\end{table*}

\newpage
\section{FAIIR Expert Assessment Survey Interface}\label{secA4}

\begin{figure}[!b]
\vspace{-2mm}
\captionsetup[subfigure]{labelformat=empty}
\begin{subfigure}[b]{0.7\textwidth}
\centering
\includegraphics[width=.8\textwidth]{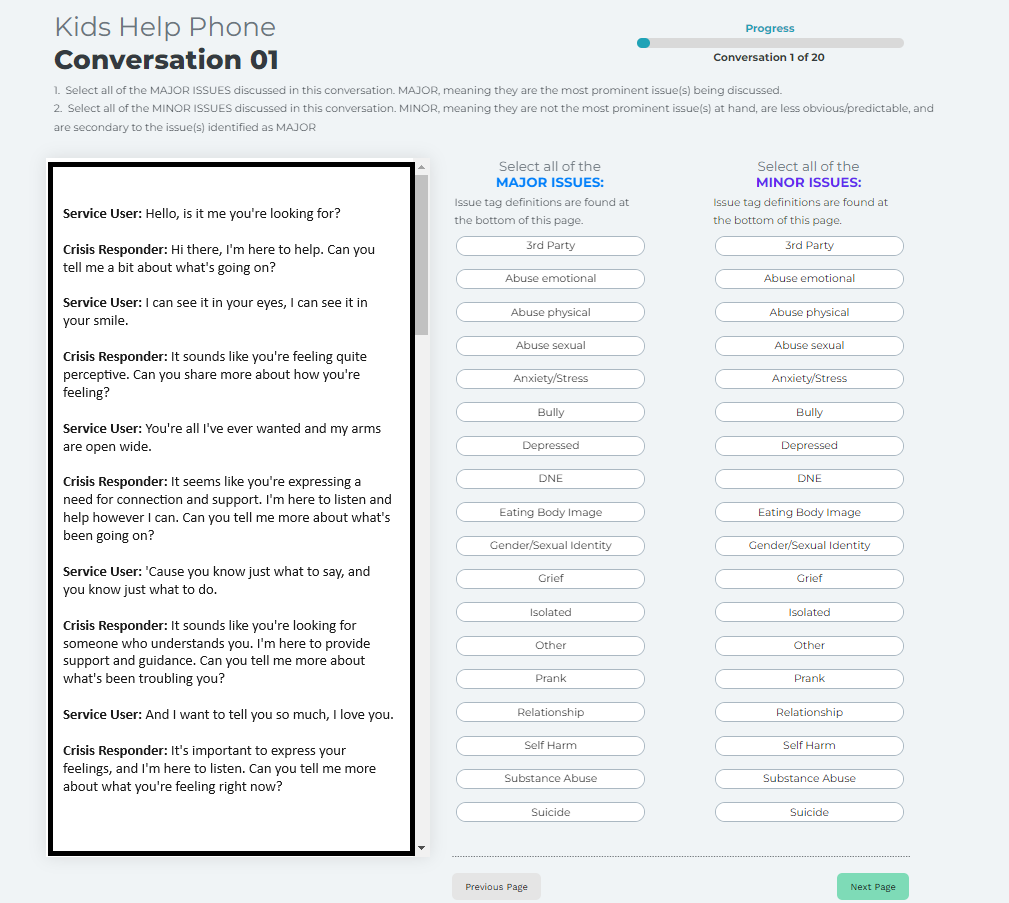}%
\caption{A)}
\end{subfigure}

\begin{subfigure}[b]{0.7\textwidth}
\centering
\includegraphics[width=.8\textwidth]{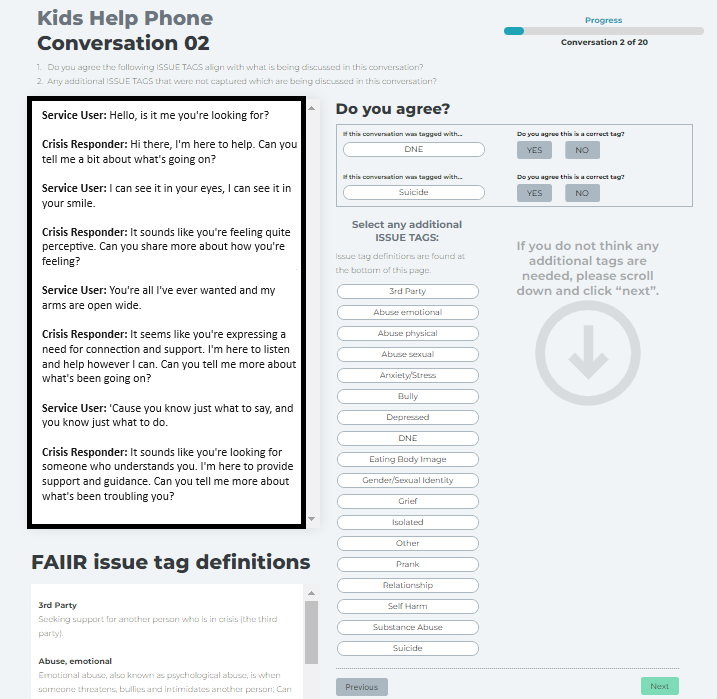}%
\caption{B)}
\end{subfigure}

\caption{Screenshot of first (A) and second (B) interface for the survey presented to experts to evaluate each conversation.} 
\label{fig:SurveyScreenshot}
\end{figure}

\newpage

\section{Silent Testing Label Distribution}
\label{appendix::silenttestingtable}
\begin{figure}[!htbp]
\centering
\includegraphics[width=.99\textwidth]{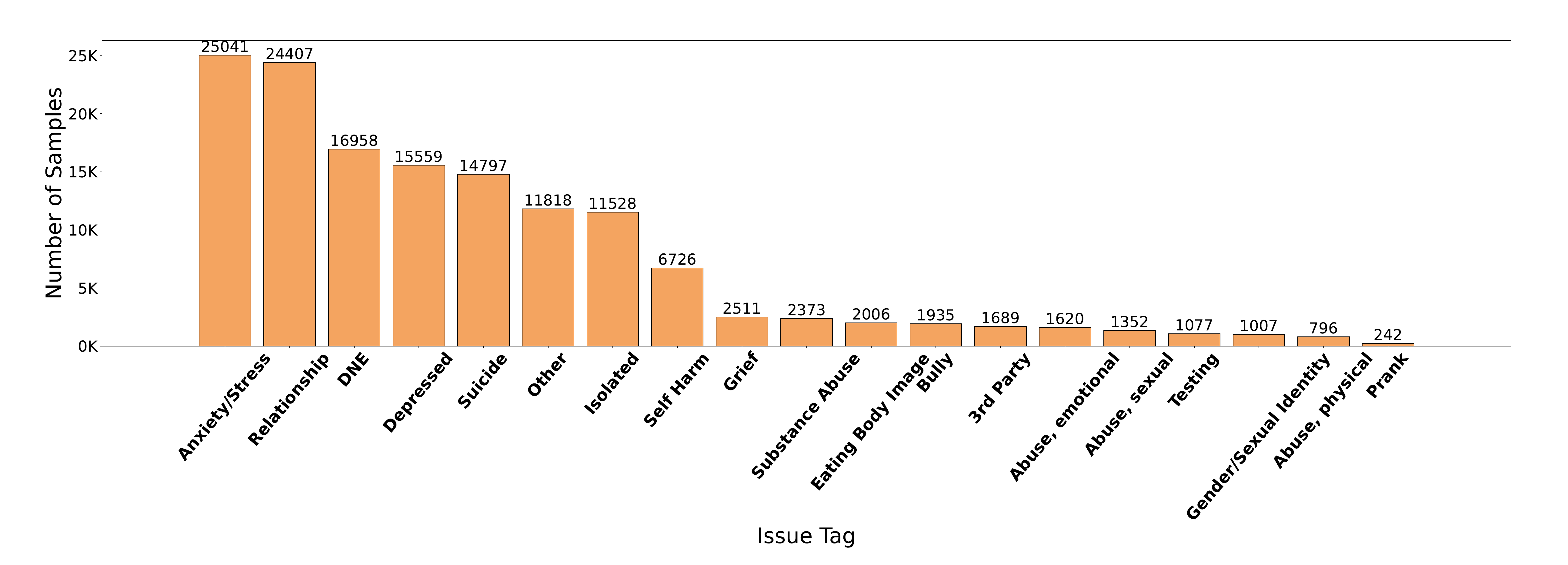}\hspace{0.2em}%
\caption{Distribution of issue tags among the set of 84,932 conversations used for silent testing. }\label{figure:silenttestingdistribution}
\end{figure}

\section{Demographic Subgroup Distributions}

\noindent\begin{minipage}{\linewidth}
\centering
\captionof{table}{Sub-group distributions as a percentage of total counts in respective demographic category. }\resizebox{\linewidth}{!}{%
\begin{tabular}{lcccccccc} 
\hline
\textbf{Type} & \multicolumn{8}{c}{\textbf{ Categories \#(\%)}} \\ 
\hline
\textbf{\textit{Gender}} & \textit{Male} & \textit{Female} & \textit{Trans Male} & \textit{Trans Female~} & \textit{Non-binary~} & \textit{Agender~} &  &  \\
 & 15.4 & 75.6 & 2.3 & 0.4 & 5.8 & 0.5 &  &  \\ 
\hline
\textbf{\textit{Orientation~}} & \textit{Heterosexual} & \textit{Gay or Lesbian} & \textit{Bisexual} & \textit{Asexual} & \textit{Unsure} &  &  &  \\
 & 55.5 & 6.5 & 27.0 & 3.3 & 7.8 &  &  &  \\ 
\hline
\textbf{\textit{Identity}} & \textit{Canadian Culture} & \textit{Disabled} & \textit{Refugee~} & \textit{Spiritual~} & \textit{Deaf~} & \textit{First Nations~} & \textit{Invisible Disability} & \textit{Other~} \\
 & 16.9 & 1.2 & 4.6 & 5.7 & 0.9 & 0.8 & 67.3 & 2.7 \\ 
\hline
\textbf{\textit{Ethnicity~}} & \textit{European Ancestry} & \textit{African or Carribean~} & \textit{Indigenous~} & \textit{East or South-East Asian~} & \textit{Middle Eastern~} & \textit{Latin American~} & \textit{South Asian~} & \textit{Unspecified} \\
 & 78.1 & 4.2 & 2.1 & 6.3 & 2.2 & 2.7 & 2.8 & 1.7 \\
\hline
\end{tabular}
}
\end{minipage}

\section{Visualization of Natural Key Words to Support Explainability}
\begin{figure*}[!htbp]
\centering
\includegraphics[width=.99\textwidth]{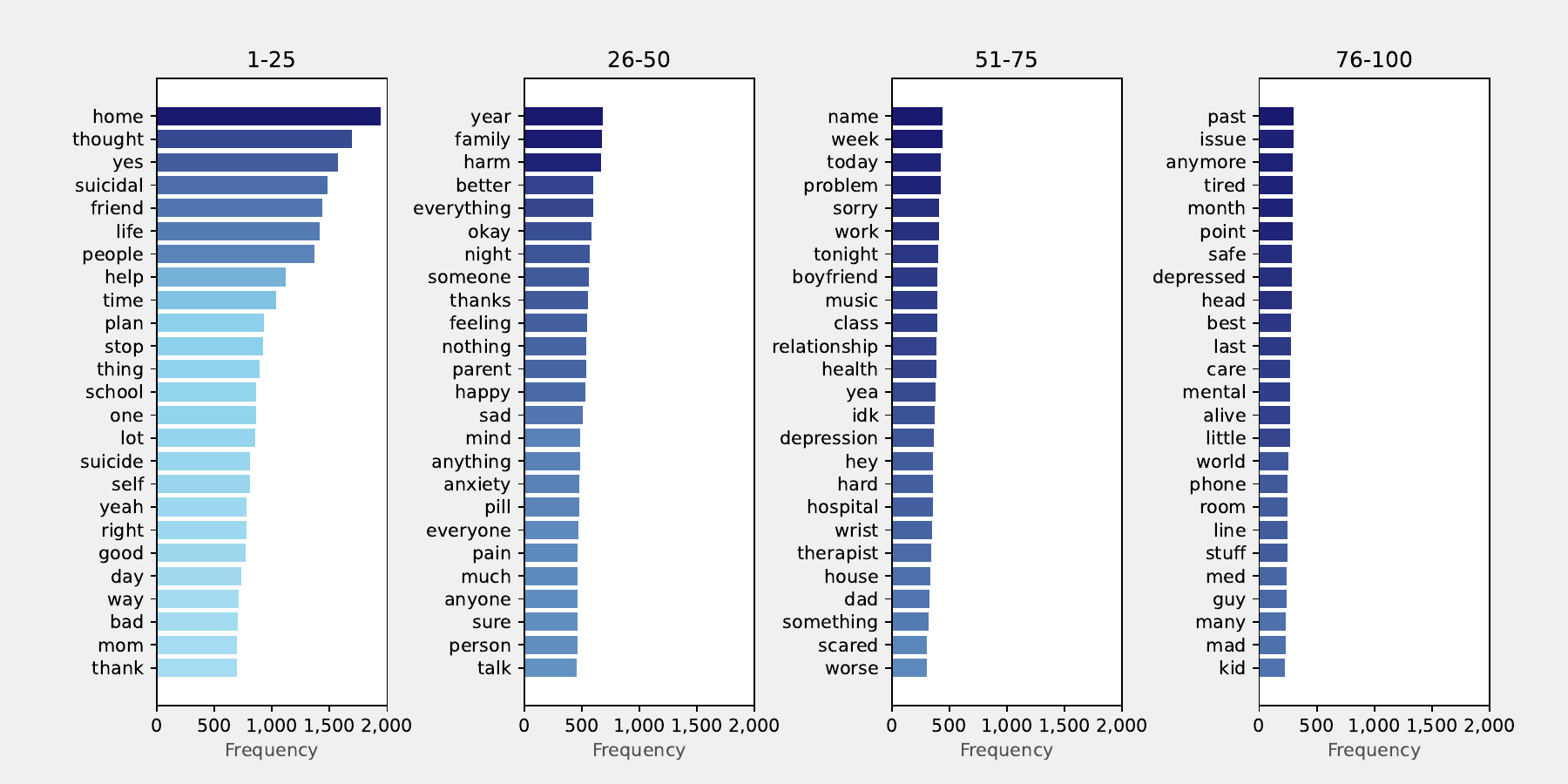}\hspace{0.2em}%
\caption{Top 100 natural keywords extracted from 10,000 conversations labelled with the \textit{Suicide} tag. } 
\label{fig:keyword-suicide}
\end{figure*}

In addition to the core 19 issue tags, we built an explainability pipeline to enable the extraction of keywords, referred to as “natural keywords”, from each conversation. Keywords are dynamic and context-specific tokens associated with the main issue tags being discussed in a conversation. By extracting these, we can derive further insights on more fine-grained important "sub-topics", that can help to better inform issues of importance. Figure \ref{fig:keyword-suicide} illustrates the occurrence frequency of the top 100 keywords across 10,000 randomly selected conversations in the test set labelled with the \textit{Suicide} issue tag. In total, 124,578 keywords were generated respectively for the \textit{Suicide} issue tag, yielding an average of 12.5 keywords per conversation. The top list of keywords generated generally represents ideas and concepts associated with the given issue tag, offering additional insights. For example, we note how in the case of suicide, frequent topics are highlighted based on common keywords, such as emotion-related keywords (i.e. "happy", "sad", "mood", "anxiety", "scared", and "pain") thus demonstrating the distress that is being experienced. Other top keywords like "plan" being in the top 10 may show that many people texting have serious plans of suicide. Location-based words like "home", "school", and "friend" also rank high, as CRs are likely trying to determine the location of the individual to try to give the most appropriate support. These insights allow for many observations at the macro-level. Similarly, Figure \ref{fig:keyword-abuse} in the appendix demonstrates the distribution of top 25 keywords across the three different abuse issue tags (\textit{Abuse, Physical}; \textit{Abuse, Sexual}; and \textit{Abuse, Emotional}), showing the general similarities of keywords across different categories (like "friend") versus those that are far more frequent in certain tags (like "assault" is to the \textit{Abuse, Sexual} tag) or how "mom" and "dad" are relatively much more frequent for \textit{Abuse, Phyiscal} and \textit{Abuse, Emotional} tags compared to the \textit{Suicide} and \textit{Abuse, Sexual} tags. Further details can be seen in Appendix \ref{secA2}

\begin{figure*}[!htbp]
\centering
\includegraphics[width=.99\textwidth]{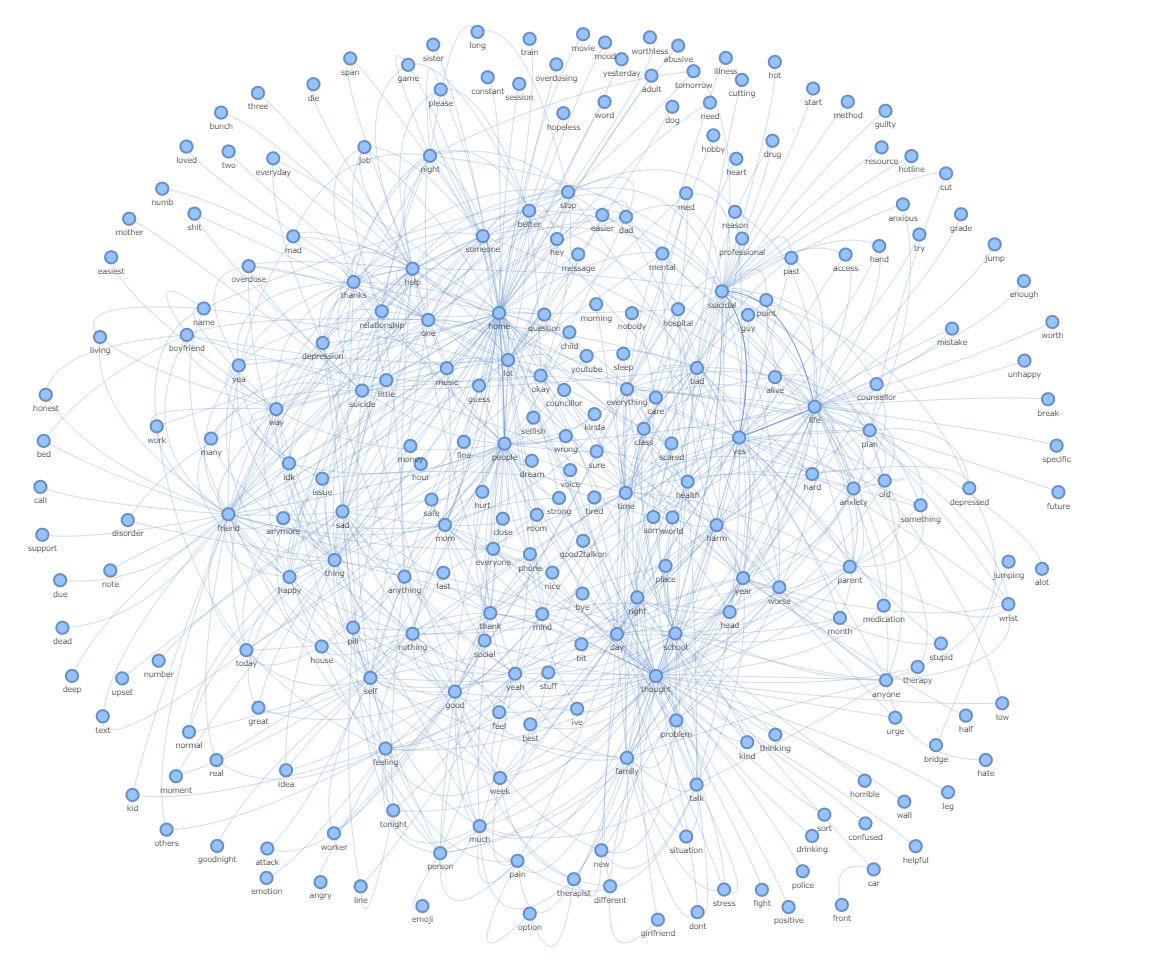}\hspace{0.2em}%
\includegraphics[width=.99\textwidth]{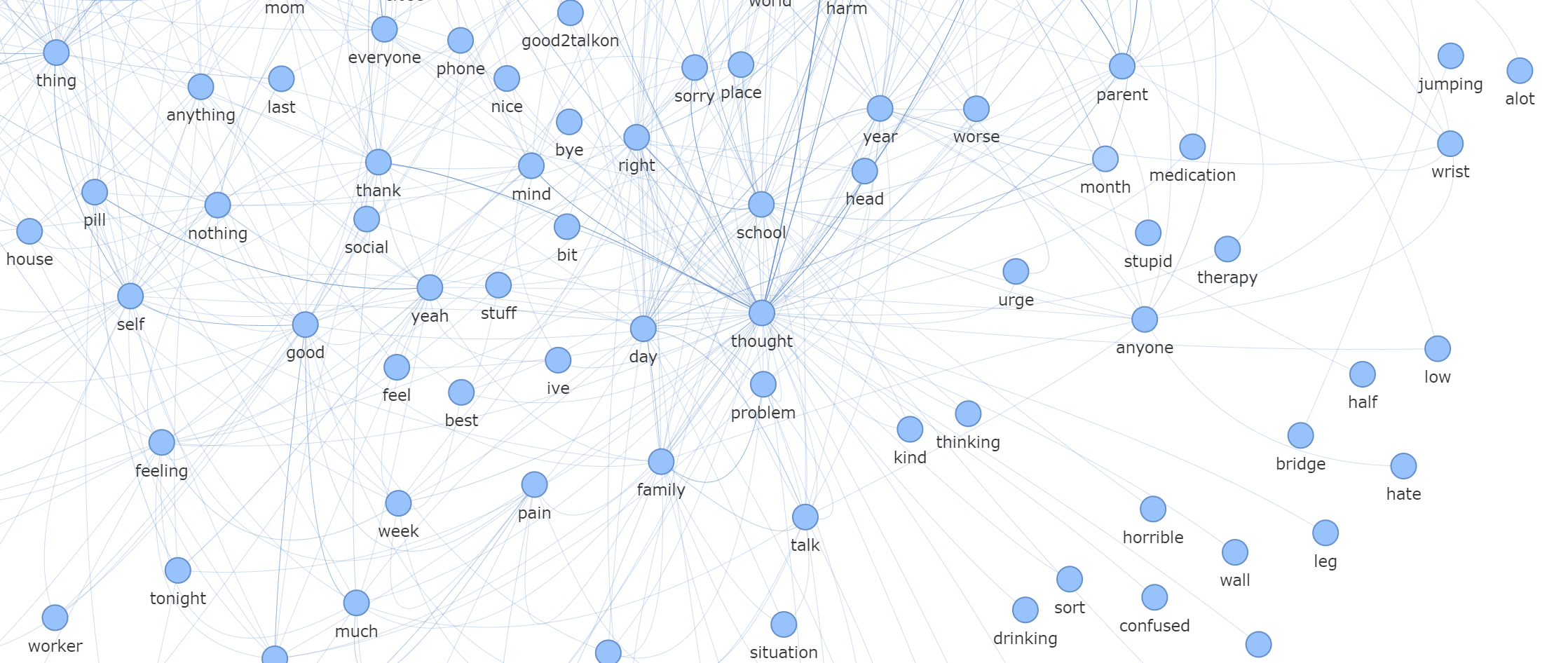}\hspace{0.2em}%
\caption{Graph of keyword bi-grams that have a co-occurrence frequency of 10 or above for the \textit{Suicide} tag. (Above) Full graph. (Below) Zoomed-In Section} 
\label{fig:of-suicide}

\end{figure*}

The explainability pipeline in the FAIIR tool offers visualization features that provide valuable insights into the semantic relationship and proximity of keywords through the utilization of bi-gram analysis and word embeddings. For bi-gram analysis, FAIIR provides a graph-based visualization in which nodes represent the keywords, while the edges between nodes illustrate the strength of their relationships (the co-occurrence frequency in conversations). Figure \ref{fig:of-suicide}, as an example, illustrates the outcome of bi-gram analysis performed on the natural keywords in relation to the \textit{suicide} issue tag (the bi-gram analysis on the \textit{Abuse, Physical} issue tag is shown in the appendix Figure \ref{fig:of-suicide}). Based on the figure, certain keywords like "thought", "suicidal", "home", and "harm" have many connections that are very strong, showing they co-occur more frequently than other pairs. Connections can reveal potential insights about behaviours or where common issues lie, which can be seen with connections like "family" and "pain", "problem", "talk", and "situation", which can potentially reveal that family troubles are frequent sub-issues tied with being suicidal.

\begin{figure*}[!htb]
\centering
\includegraphics[width=.99\textwidth]{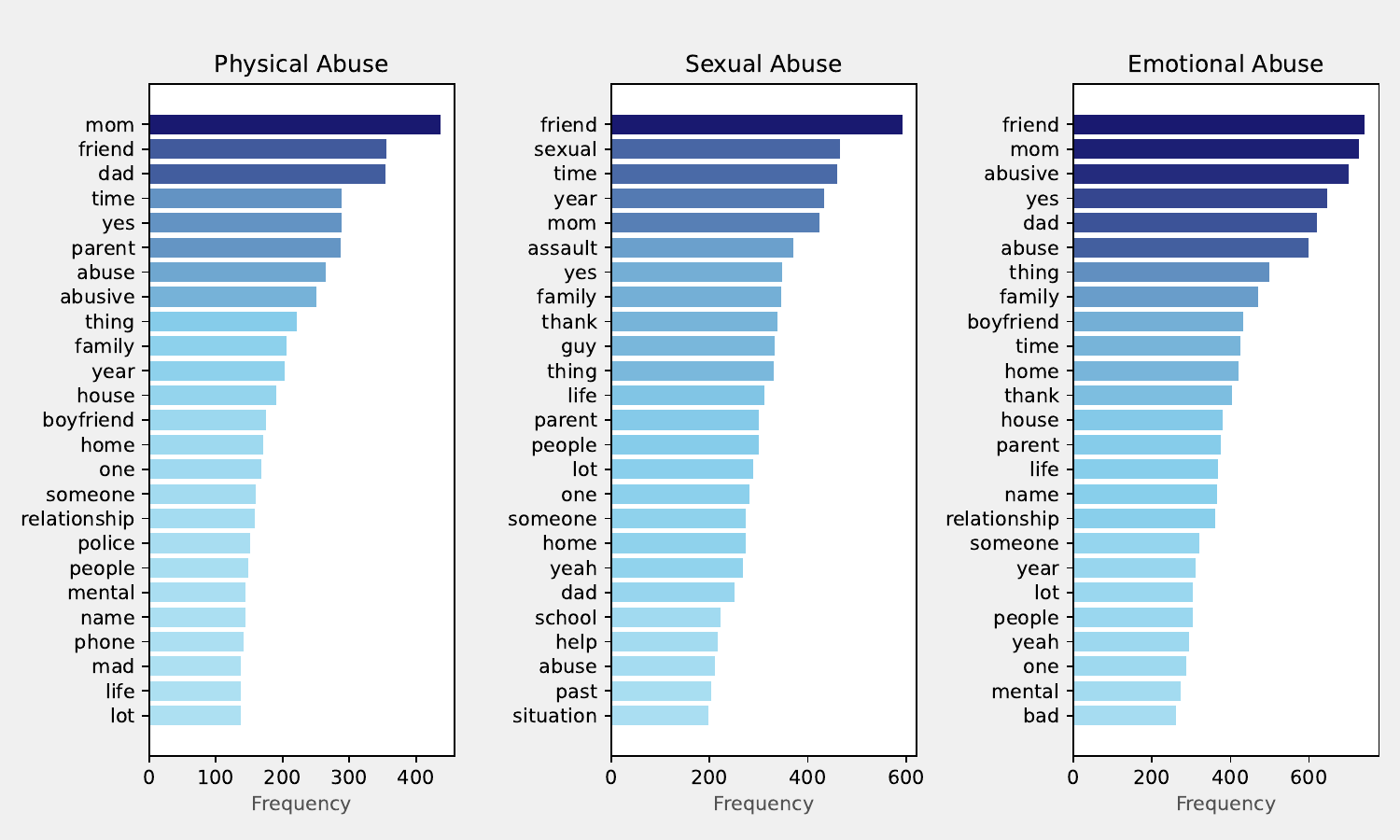}\hspace{0.2em}%
\caption{Top 25 natural keywords extracted from 1673 conversations labelled with \textit{Abuse, Physical}; 2598 conversations with \textit{Abuse, Sexual}; and 3367 conversations with \textit{Abuse, Emotional}). Top 25 natural keywords extracted from 1673 conversations labelled with \textit{Abuse, Physical}; 2598 conversations with \textit{Abuse, Sexual}; and 3367 conversations with \textit{Abuse, Emotional}). } 
\label{fig:keyword-abuse}

\end{figure*}

\begin{figure*}[!htbp]
\centering
\includegraphics[width=.99\textwidth]{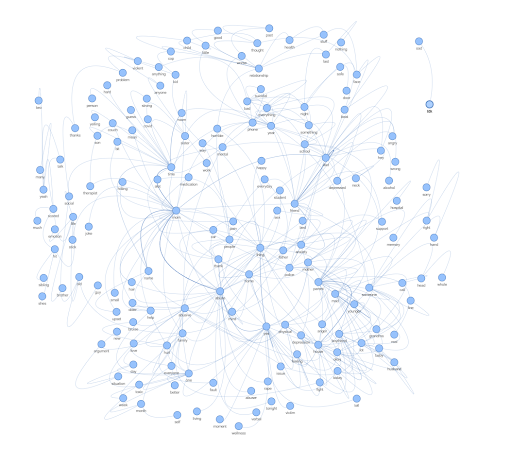}\hspace{0.2em}%
\includegraphics[width=.9\textwidth]{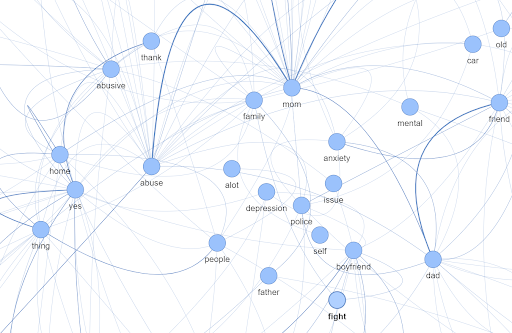}\hspace{0.2em}%
\caption{Graph of keyword bi-grams that have a co-occurrence frequency of 5 or above for the \textit{Abuse, Physical} tag} 
\label{fig:of-abuse}

\end{figure*}

\subsection{Further Insights on Extracted Key Words}\label{secA2}

Figure \ref{fig:of-suicide} illustrates the occurrence frequency of the top 25 keywords across all conversations in the test set labelled with the three abuse issue tags (\textit{Abuse, Physical}; \textit{Abuse, Sexual}; and \textit{Abuse, Emotional}). In total, 1,673, 2,598, and 3,367 conversations were used to generate 26,985, 40,522 and 52,635 keywords, respectively, for the three aforementioned issue tags, yielding an average of 16.2, 15.6, and 15.6 keywords per conversation. The top list of keywords generated generally represents ideas and concepts associated with the given issue tag, offering additional insights. For example, we note how in the case of sexual abuse, both "home" and "school" are frequent keywords, with home being relatively more frequent than school. This may reflect places where service users are more likely to have experienced abuse, thereby aiding in strategy and planning for CRs in managing conversations. We observe similar keywords are common across these issue tags, such as "friend", "mom", or "dad" which reveal major overarching topics and pressure points.

\begin{figure*}[!htbp]
\centering
\includegraphics[width=.99\textwidth]{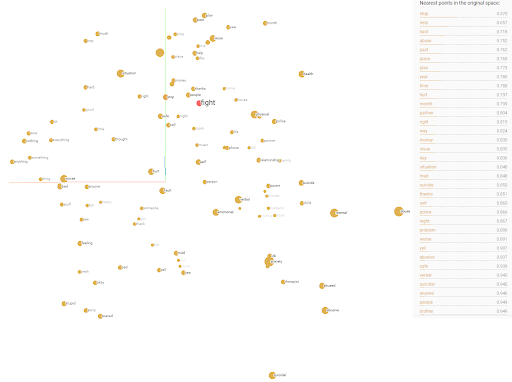}\hspace{0.2em}%
\vspace{-2mm}
\caption{Using GloVe embeddings on the top 100 keywords closest to the issue tag \textit{Abuse, Emotional}, centering on the word ``fight''.  PCA is then employed to reduce the dimensionality of the original 50-dimensional GloVe word vectors, enhancing the visualization of the semantic relationships between these keywords.} 
\label{fig:tensorboard_figure}
\end{figure*}

In addition to bi-gram analysis, in Figure \ref{fig:tensorboard_figure} we demonstrate the word embeddings of the top 100 most frequent keywords within all conversations originally labelled with the issue tag \textit{Abuse, Emotional}, projected in a three-dimensional space using Principal Component Analysis (PCA). This enables enhanced visualization and easier exploration of keywords that are most similar to a searched keyword using cosine distance between the embeddings. For example, for the keyword "fight", keywords like "stop", "help" and "hard" are the closest in embedding space.

\newpage

\end{appendices}

\bibliography{sn-bibliography}

\end{document}